\documentclass{article}

\usepackage[final]{neurips_2025}

\usepackage[utf8]{inputenc} 
\usepackage[T1]{fontenc}    
\usepackage{hyperref}       
\usepackage{url}            
\usepackage{booktabs}       
\usepackage{amsfonts}       
\usepackage{nicefrac}       
\usepackage{microtype}      
\usepackage{xcolor}         
\usepackage{subcaption}
\usepackage{graphicx}
\usepackage{amsmath}
\usepackage{wrapfig} 
\usepackage{multirow}
\usepackage{algorithm}
\usepackage{algorithmic}

\title{BioOSS: A Bio-Inspired Oscillatory State System with Spatio-Temporal Dynamics}

\author{
  Zhongju Yuan$^{1}$\thanks{Corresponding author.} \quad
  Geraint Wiggins$^{2,3}$ \quad
  Dick Botteldooren$^{1}$  \\
  $^{1}$WAVES Research Group, Ghent University, Gent, Belgium \\
  $^{2}$AI Lab, Vrije Universiteit Brussel, Brussel, Belgium \\
  $^{3}$EECS, Queen Mary University of London, London, UK \\
  \texttt{zhongju.yuan@ugent.be}, \quad
  \texttt{geraint.wiggins@vub.be}, \quad
  \texttt{dick.botteldooren@ugent.be} 
}

\begin{document}

\maketitle

\begin{abstract}
Today’s deep learning architectures are primarily based on perceptron models, which do not capture the oscillatory dynamics characteristic of biological neurons. Although oscillatory systems have recently gained attention for their closer resemblance to neural behavior, they still fall short of modeling the intricate spatio-temporal interactions observed in natural neural circuits. In this paper, we propose a \textbf{bio}-inspired \textbf{o}scillatory \textbf{s}tate \textbf{s}ystem (BioOSS) designed to emulate the wave-like propagation dynamics critical to neural processing, particularly in the prefrontal cortex (PFC), where complex activity patterns emerge. BioOSS comprises two interacting populations of neurons: \(p\) neurons, which represent simplified membrane-potential-like units inspired by pyramidal cells in cortical columns, and \(o\) neurons, which govern propagation velocities and modulate the lateral spread of activity. Through local interactions, these neurons produce wave-like propagation patterns. The model incorporates trainable parameters for damping and propagation speed, enabling flexible adaptation to task-specific spatio-temporal structures. We evaluate BioOSS on both synthetic and real-world tasks, demonstrating superior performance and enhanced interpretability compared to alternative architectures.
\end{abstract}

\section{Introduction}

The advent of deep learning models, such as transformers~\citep{vaswani2017attention, devlin2019bert}, has significantly advanced performance across a variety of tasks and domains. However, these models are still fundamentally based on perceptrons, which differ from the oscillatory behavior of biological neurons. In contrast, oscillatory systems~\citep{rusch2021coupled, lanthaler2023neural, rusch2025oscillatory} have recently garnered considerable attention due to their closer alignment with the dynamic properties of biological neurons. While these models capture certain temporal aspects of neural oscillations, they fail to replicate the complex spatio-temporal interplay characteristic of biological neural circuits. This limitation arises because biological oscillations are typically driven by coupled neurons with both temporal and spatial interactions. Consequently, existing oscillatory models primarily capture temporal dependencies; even when spatial couplings are included, they do not account for the structured spatial organization and physical distances inherent in real neural circuits.

\begin{figure}[t]
  \centering
  \includegraphics[width=0.95\linewidth]{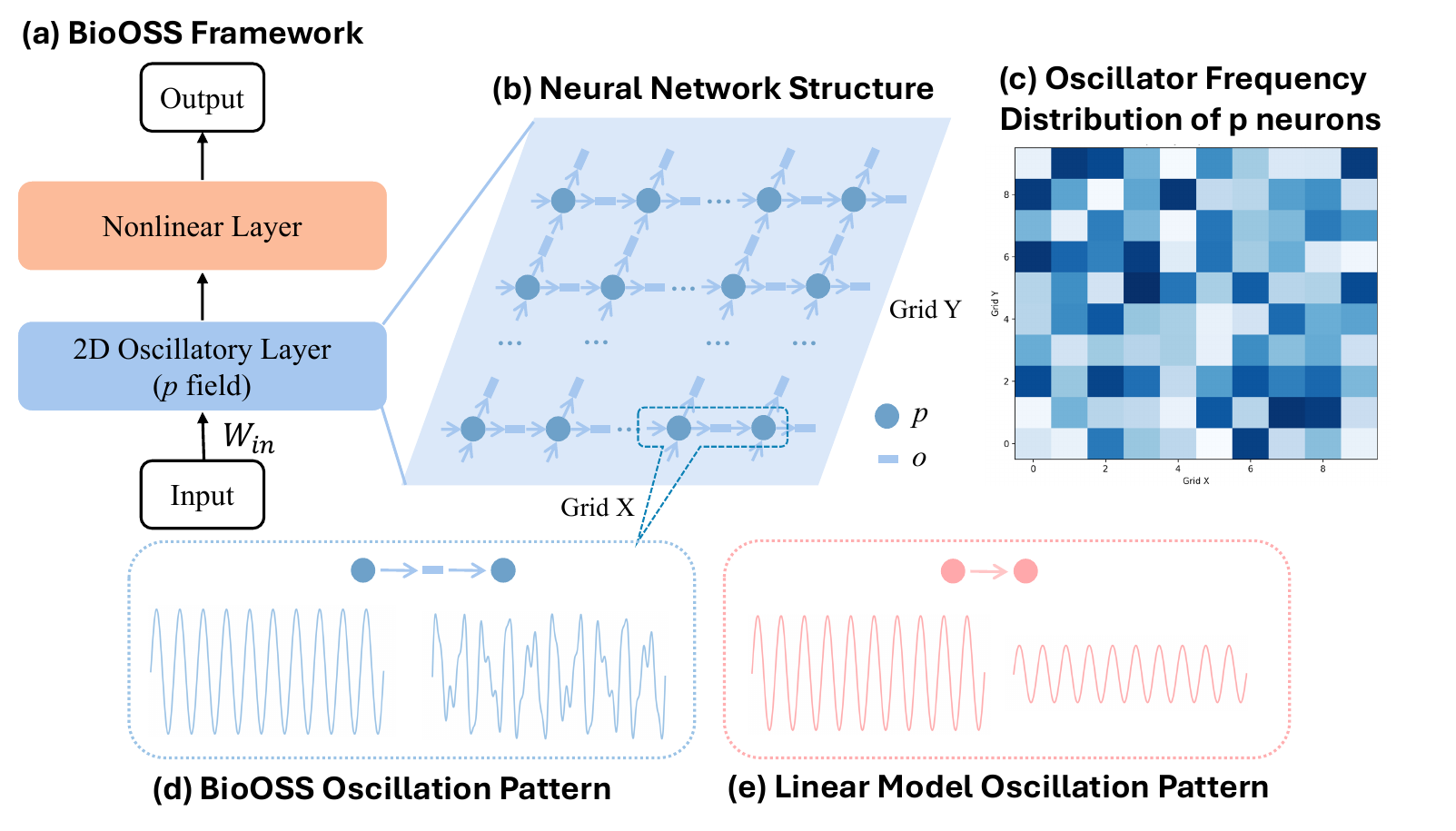}
  \caption{\textbf{(a)} The Overview of the proposed BioOSS framework.
\textbf{(b)} Structure of the 2D neural network composed of interacting $p$ (pressure-like) and $o$ (oscillation-like) neurons arranged on a spatial grid.
\textbf{(c)} Distribution of the natural frequencies across the grid, illustrating local frequency heterogeneity.
\textbf{(d)} Dynamic behavior of \(p\) neurons in the proposed model with local interactions. 
The left panel shows the input signal to a left \(p\) neuron; the right panel shows the enriched oscillatory pattern of a neighboring right \(p\) neuron.
\textbf{(e)} Behavior in the absence of local spatial interactions, where the right \(p\) neuron exhibits only amplitude modulation without generating richer oscillatory dynamics.
}
\label{fig:overview}
\end{figure}

We propose a \textbf{bio}-inspired \textbf{o}scillatory \textbf{s}tate \textbf{s}ystem (BioOSS) to model the spatio-temporal wave dynamics observed in neural circuits. In the brain, neural states evolve not only over time but also through spatial interactions with neighboring neurons, forming the basis for neural dynamic computation. Such dynamics are particularly salient in the prefrontal cortex (PFC), where complex and coordinated activity patterns support higher-order cognitive functions~\citep{dale2001spatiotemporal, buonomano2009state}. Temporal recurrence and delay mechanisms enable memory and integration~\citep{fuster2012cognit, richards2017persistence}, while spatially structured cortical columns facilitate localized wave propagation~\citep{watakabe2023local}. BioOSS captures these properties in a unified framework (Fig.~\ref{fig:overview}a), offering a biologically grounded approach to modeling neural dynamics and enabling the development of more generalizable and interpretable artificial systems.

As observed in \citep{watakabe2023local}, neural circuits tend to organize primary signal-carrying units into structured rows rather than random patterns. To capture this property, we propose a 2D grid-based framework composed of two interacting neuronal populations: \(p\) and \(o\) neurons (Fig.~\ref{fig:overview}b). \(p\) neurons serve as the main signal carriers in our model, functionally inspired by pyramidal neurons in cortical columns as described in~\citep{watakabe2023local}, and form a distributed oscillatory pattern with diverse frequencies (Fig.~\ref{fig:overview}c). In parallel, \(o\) neurons represent propagation velocities, mimicking diffuse projections that modulate activity across cortical tissue. The local interactions between neurons further enrich the oscillatory dynamics, giving rise to complex patterns beyond simple harmonic components (Fig.~\ref{fig:overview}d). In contrast, purely linear transformations, such as fully connected layers without non-linearity, only reweight and shift existing components without introducing new frequency content (Fig.~\ref{fig:overview}e).

The \(p\) and \(o\) neurons are locally interconnected, enabling wave-like lateral signal propagation. To introduce biologically plausible flexibility, we incorporate trainable parameters, including damping \(k\) and propagation speed \(c\), which reflect region-specific attenuation and projection density~\citep{alcami2019beyond, campagnola2022local}. By learning these parameters from data, the model can adaptively generate task-relevant spatio-temporal patterns. Furthermore, the spatio-temporal couplings allow us to derive connectome eigenmodes for each grid~\citep{xia2024decomposing}, enriching the oscillatory dynamics and improving performance on time series tasks.

In summary, our contributions are:
\begin{itemize}
    \item We introduce BioOSS, a biologically inspired oscillatory system that models simplified spatio-temporal wave propagation, drawing analogies to cortical column dynamics without aiming for full biological realism.
    \item We design an efficient trainable framework with flexible damping and propagation speed parameters, enabling adaptive and stable spatio-temporal pattern generation across tasks.
    \item We validate BioOSS through extensive experiments on synthetic and real-world datasets, demonstrating both competitive performance and enhanced interpretability compared to baseline architectures.
\end{itemize}

\section{Related Work}

\paragraph{Neuroscience Inspiration.} 
Our model is inspired by neuroscience findings that cognitive functions and information processing emerge from synchronized oscillations across coupled neuronal populations. These neural oscillations, arising from the interaction between excitatory and inhibitory processes, exhibit rhythmic, coordinated activity patterns~\citep{doelling2021neural, esghaei2022dynamic}. Recent studies~\citep{watakabe2023local} show that neurons in the prefrontal cortex (PFC) are organized in a discrete, mosaic-like structure, where signal-carrying units communicate diffusely via intermediate neurons. This spatial organization enables circuits to generate complex oscillatory dynamics from multi-dimensional interactions. Building on this biological insight, we design a simplified 2D neural network composed of a grid of primary neurons coupled with intermediate neurons. By training the damping coefficients and propagation speeds, our model simulates wave-like dynamics~\citep{foster2024brain} and resonance phenomena~\citep{rigotti2013importance} observed in biological circuits.

\paragraph{Comparison to Existing Sequence Modeling Approaches.} 
Sequence modeling is central to capturing spatio-temporal dependencies in both biological and artificial systems. Transformers~\citep{wen2023transformers} capture temporal relationships through attention, RNNs~\citep{torres2021deep} maintain hidden states across time, and State Space Models (SSMs)~\citep{auger2021guide} describe latent state evolution. 

Oscillatory extensions move beyond perceptron-like updates. LinOSS~\citep{rusch2025oscillatory} exemplifies this by introducing second-order ODE dynamics to model rhythmic activity within scan-based architectures, while providing stability guarantees. BioOSS builds on this line of work but diverges in key ways: it employs a 2D neural grid with local couplings instead of 1D hidden states, grounds its recurrence in first-order PDEs inspired by wave propagation rather than ODEs, and incorporates distinct signal-bearing and modulatory units to yield spatially structured, biologically interpretable dynamics. These design choices result in different numerical properties and spatio-temporal representations.

Beyond LinOSS, recent works have explored oscillatory and wave-like computation using varied abstractions. Neural Wave Machines~\citep{keller2023neural} employ convolutional couplings on a topographic grid, spectral approaches~\citep{balkenhol2024beyond} use graph filters, and analog substrates such as Wave-RNNs~\citep{hughes2019wave} or water-wave reservoirs~\citep{maksymov2023analogue} approximate physical propagation. These methods highlight the growing interest in wave-based computation, but approach it through different mechanisms, such as matrix couplings, graph filters, analog substrates, or through PDE formulations with different goals (e.g., spatial convolutional operators), rather than a discretized cortical-inspired PDE system as in BioOSS.

BioOSS extends this landscape by introducing a wave-PDE-driven, discretized dynamical system compatible with scan operations. Its analytical eigendecomposition yields efficient Fourier-domain recurrence, enabling closed-form characterization of oscillatory modes and frequency responses. This combination of spatial structure, biological grounding, and computational tractability distinguishes BioOSS from both LinOSS and other oscillatory models, while maintaining interoperability with structured SSM frameworks.

\section{Methods}
\subsection{Overview of the proposed model}
We propose BioOSS, illustrated in Fig.~\ref{fig:overview}a, whose core component is a two-dimensional oscillatory layer (Fig.~\ref{fig:overview}b). This biologically motivated model consists of two neuron types: \(p\) neurons, which serve as the primary computational units and carry the signal, and \(o\) neurons, which modulate the velocity of information propagation. The spatio-temporal dynamics of the system are described by the following equations:
\begin{align}
    \frac{\partial p}{\partial t} + k^p p + c^2 \nabla \cdot \mathbf{o} &= \mathbf{B} \mathbf{u}(t), \\
    \frac{\partial \mathbf{o}}{\partial t} + \mathbf{k}^{o} \mathbf{o} + \nabla p &= 0,
\label{eq:main_equations}
\end{align}
where \( p \) denotes the \( p \) neuron value, \( \mathbf{o} = (o_x, o_y) \) represents the \( o \) neuron values in the \( x \) and \( y \) directions, \( k^{p} \) and \( \mathbf{k}^{o} \) are the damping coefficients, \( c \) is the wave speed, and \( \mathbf{u}(t) \) is the external input at time step \( t \), while \(\mathbf{B} \) is the input weight matrix. For convenience, the system can be expressed in matrix form as:
\begin{equation}
    \mathbf{x}'(t) =
    \begin{bmatrix}
        - k^p & -c^2 \nabla \cdot \\
        -\nabla & -\mathbf{k}^{o}
    \end{bmatrix}
    \mathbf{x}(t)
    +
    \begin{bmatrix}
        \mathbf{B} \\
        0
    \end{bmatrix}
    \mathbf{u}(t) = \mathbf{A} \mathbf{x}(t) + \begin{bmatrix}
        \mathbf{B} \\
        0
    \end{bmatrix}
    \mathbf{u}(t),
\label{eq:hidden_state_PDE}
\end{equation}
where \( \mathbf{x}(t) = 
    \begin{bmatrix} 
        p,  
        \mathbf{o} 
    \end{bmatrix}^{T} \) and \(\mathbf{A}\) is the coupling matrix.

The output equation is given by:
\begin{equation}
    \mathbf{y}(t) = \mathbf{C} \mathbf{x}(t) + \mathbf{D} \mathbf{u}(t),
\end{equation}
where \( \mathbf{C} \) and \( \mathbf{D} \) are the linear weight matrices.
\subsection{Discretization Scheme}
To mitigate the significant computational cost associated with solving the PDE at each time step, we propose an explicit discretization scheme that generates an explicit state transition matrix. This approach avoids the need for a solver, enabling more efficient computation.

We begin by updating the wave propagation with a fixed timestep \(\Delta t\) as follows:
\begin{align}
    \mathbf{o}_n^* &= \mathbf{o}_{n-1} - \Delta t \nabla p_{n-1}, \\
    p_n^* &= p_{n-1} - c^2 \Delta t \nabla \cdot \mathbf{o}_n^* + \Delta t \mathbf{B} \mathbf{u}_n.
\end{align}

Next, we apply a damping correction as follows:
\begin{align}
    (\mathbf{I} + \Delta t \mathbf{k}^{o}) \mathbf{o}_n &= \mathbf{o}_n^*, \\
    (1 + \Delta t k^p) p_n &= p_n^*.
\end{align}

The full update equation is then given by:
\begin{equation}
    \mathbf{x}_n = \mathbf{M}^{\text{Damp}^{-1}} \mathbf{M}^{\text{Velocity}} \mathbf{x}_{n-1} + \mathbf{M}^{\text{Damp}^{-1}} \mathbf{F}^{\text{Velocity}}_n = \mathbf{A} \mathbf{x}_{n-1} + \mathbf{M}^{\text{Damp}^{-1}} \mathbf{F}^{\text{Velocity}}_n,
\label{eq:update_equation}
\end{equation}
where
\(
\mathbf{M}^{\text{Velocity}} =
    \begin{bmatrix}
        1 & -c^2 \Delta t \nabla \cdot \\
        - \Delta t \nabla & \mathbf{I}
    \end{bmatrix}, 
\)
\(
\mathbf{M}^{\text{Damp}} =
    \begin{bmatrix}
        1 + \Delta t k^p & 0 \\
        0 & \mathbf{I} + \Delta t \mathbf{k}^{o}
    \end{bmatrix},
\)
and
\(
\mathbf{F}^{\text{Velocity}}_n =
    \begin{bmatrix}
        \Delta t \mathbf{B} \mathbf{u}_n \\
        0
    \end{bmatrix}.
\)

Although \(\mathbf{A} = \mathbf{M}^{\text{Damp}^{-1}} \mathbf{M}^{\text{Velocity}}\) serves as a coupling matrix to capture internal propagation. 


Since $\mathbf{M}^{\text{Damp}}$ is diagonal by construction, its inverse is obtained by taking the reciprocal of its diagonal entries:
\begin{equation}
\mathbf{M}^{\text{Damp}^{-1}} = 
\operatorname{diag}\!\left(\frac{1}{1 + \Delta t\, k^{p}_{i,j}}, \frac{1}{1 + \Delta t\, k^{o}_{i,j}}\right),
\end{equation}
where each diagonal entry is simply the reciprocal of the corresponding factor $(1 + \Delta t\, k^{p}_{i,j})$ or $(1 + \Delta t\, k^{o}_{i,j})$ at grid point $(i,j)$. 

Therefore, the coupling matrix becomes the following form:
\begin{align}
\mathbf{A} = \mathbf{M}^{\text{Damp}^{-1}}\mathbf{M}^{\text{Velocity}} = 
\begin{pmatrix}
(\mathbf{I}+\Delta t\mathbf{k}^p)^{-1} & -\mathbf{c}^2\Delta t(\mathbf{I}+\Delta t\mathbf{k}^o)^{-1}\nabla\cdot \\
-\Delta t(\mathbf{I}+\Delta t\mathbf{k}^{p})^{-1}\nabla & (\mathbf{I}+\Delta t\mathbf{k}^{o})^{-1}
\end{pmatrix}
\end{align}

Since both $\mathbf{c}$, $\mathbf{k}^p$ and $\mathbf{k}^{o}$ are matrices with potentially different values at each grid point, we need to perform a local analysis. For each grid point $(i,j)$, we examine the local evolution of the system.

\subsection{Efficient Recurrent Scan Operator}

The computational efficiency of BioOSS is limited because the recurrence is inherently sequential and cannot be parallelized, while the spatial operators (gradient and divergence) must be applied at every time step. 
To alleviate this cost, we adopt the scan operator together with an eigendecomposition-based approximation of the system matrix, which reduces the burden of repeatedly applying the spatial operators. 

The main challenge is that the recurrence induced by these operators breaks the associativity property 
\((x \bullet y) \bullet z = x \bullet (y \bullet z)\)~\citep{rusch2025oscillatory}. 
Here, \(x, y, z\) denote intermediate system states, typically represented as tuples \((\mathbf{A}, \mathbf{M}^{\text{Damp}^{-1}} \mathbf{F}^{\text{Velocity}})\) combining the system matrix and the input vector at each step, and \(\bullet\) is a binary operation such as 
\((\mathbf{a}_1, \mathbf{a}_2) \bullet (\mathbf{b}_1, \mathbf{b}_2) = (\mathbf{b}_1 \mathbf{a}_1, \mathbf{b}_1 \mathbf{a}_2 + \mathbf{b}_2)\)~\citep{rusch2025oscillatory}. 
As emphasized in recent work on oscillatory state space models~\citep{rusch2025oscillatory}, associativity is essential because it enables parallel scan operations, reducing the computational complexity of a serial recurrence from \(\mathcal{O}(N)\) to \(\mathcal{O}(\log N)\). 
Following SSM literature~\citep{orvieto2023resurrecting,rusch2025oscillatory}, we therefore apply eigendecomposition 
\(\mathbf{A} = \mathbf{P} \boldsymbol{\Lambda} \mathbf{P}^{-1}\), 
where \(\boldsymbol{\Lambda}\) is diagonal. 
This diagonalization restores associativity and allows efficient scan operations across time steps, significantly lowering the computational cost of sequential updates.

To find the eigenvalues, we transform to the Fourier domain where gradient, and divergence operators become algebraic in Fourier space, as detailed in the proof provided in the Appendix Section~\ref{appendix:Fourier transform}. For spatial frequency components $(\xi_x, \xi_y)$ and for each grid point $(i,j)$ with local values of $c_{i,j}$, $k^p_{i,j}$, and $k^{o}_{i,j}$:
\begin{align}
\nabla p &\rightarrow (i\xi_x p, i\xi_y p) \\
\nabla \cdot \mathbf{o} &\rightarrow i\xi_x o_x + i\xi_y o_y
\end{align}

For each Fourier mode $(\xi_x, \xi_y)$ and each local grid point, the system matrix becomes a $3 \times 3$ matrix:
\begin{equation}
\begin{pmatrix}
\frac{1}{1+\Delta t k^p_{i,j}} & -\frac{c^2\Delta t i\xi_x}{1+\Delta t k^p_{i,j}} & -\frac{c^2\Delta t i\xi_y}{1+\Delta tk^p_{i,j}} \\
-\frac{\Delta t i\xi_x}{1+\Delta tk^{o}_{i,j}} & \frac{1}{1+\Delta tk^{o}_{i,j}} & 0 \\
-\frac{\Delta t i\xi_y}{1+\Delta tk^{o}_{i,j}} & 0 & \frac{1}{1+\Delta tk^{o}_{i,j}}
\end{pmatrix}.
\label{eq:grid_matrix}
\end{equation}
This structure can be expanded into a coupling matrix of size $3 \times \text{grid\_size} \times 3 \times \text{grid\_size}$, capturing the spatio-temporal coupling weights among all neurons. Then, we solve the characteristic equation \( \det(\mathbf{A} - \lambda\mathbf{I}) = 0 \) to find the eigenvalues $\lambda$. By applying a cubic equation, we can get the following simplified equation:
\begin{equation}
\small
\left(\frac{1}{1+\Delta tk^o_{i,j}} - \lambda\right)\left[\left(\frac{1}{1+\Delta tk^p_{i,j}} - \lambda\right)\left(\frac{1}{1+\Delta tk^o_{i,j}} - \lambda\right) + \frac{c_{i,j}^2\Delta t^2(\xi_x^2 + \xi_y^2)}{(1+\Delta tk^p_{i,j})(1+\Delta tk^o_{i,j})}\right] = 0
\end{equation}

From the factored characteristic equation, we can identify two distinct eigenvalue solutions: a real eigenvalue corresponding to one component of the velocity field, \(\lambda_1 = \frac{1}{1+\Delta tk^o_{i,j}}\), and the other two eigenvalues are the solutions to the right part quadratic term.

Applying the quadratic formula under the assumption that the damping coefficients are approximately equal ($k^p_{i,j} \approx k^o_{i,j}$) and the spatial frequencies are sufficiently large, the discriminant becomes negative. Consequently, the eigenvalues form a pair of complex conjugates:
\begin{equation}
\lambda_{2,3} \approx \frac{1}{2}\left(\frac{1}{1+\Delta tk^p_{i,j}} + \frac{1}{1+\Delta tk^o_{i,j}}\right) \pm i\frac{c_{i,j}\Delta t\sqrt{\xi_x^2 + \xi_y^2}}{\sqrt{(1+\Delta tk^p_{i,j})(1+\Delta tk^o_{i,j})}}.
\label{eq:lambda_2_3}
\end{equation}

Given the availability of local eigenvalues in BioOSS, the corresponding eigenvectors can also be computed. The complete eigenvector matrix is then assembled as follows (details of the derivation are provided in the Appendix Section~\ref{appendix:eigenvector}):
\[
\mathbf{P} = 
\begin{bmatrix}
| & | & | \\
\mathbf{v}_1 & \mathbf{v}_2 & \mathbf{v}_3 \\
| & | & |
\end{bmatrix}
=
\begin{bmatrix}
0 & 1 & 1 \\
1 & \displaystyle \frac{\Delta t \cdot i \xi_x}{(1 + \Delta t\,k^o)(\lambda_2 - \lambda_1)} & \displaystyle \frac{-\Delta t \cdot i \xi_x}{(1 + \Delta t\,k^o)(\lambda_3 - \lambda_1)} \\
0 & \displaystyle \frac{\Delta t \cdot i \xi_y}{(1 + \Delta t\,k^o)(\lambda_2 - \lambda_1)} & \displaystyle \frac{-\Delta t \cdot i \xi_y}{(1 + \Delta t\,k^o)(\lambda_3 - \lambda_1)}
\end{bmatrix}
\in \mathbb{C}^{3 \times 3}.
\]

Leveraging the scan property and the eigendecomposition of the system matrix, we express the update operator as \( \mathbf{A} = \mathbf{P} \boldsymbol{\Lambda} \mathbf{P}^{-1} \), where \( \mathbf{P} \) contains the eigenvectors and \( \boldsymbol{\Lambda} \) is a diagonal matrix of eigenvalues. Notably, the inverse \( \mathbf{P}^{-1} \) is computed only once per sequence and backpropagation iteration during training, thereby reducing the computational overhead in iterative evaluations.

\subsection{Stability Analysis}

The system exhibits a recurrent dynamic structure similar to that in linear state-space models. 
Specifically, given the eigendecomposition of the state transition matrix 
\(\mathbf{A} = \mathbf{P}\boldsymbol{\Lambda}\mathbf{P}^{-1}\), the recurrence takes the form
\begin{equation}
\mathbf{x}_n = \mathbf{A}\mathbf{x}_{n-1} + \mathbf{B}\mathbf{u}_n 
= \mathbf{P}\boldsymbol{\Lambda}\mathbf{P}^{-1}\mathbf{x}_{n-1} + \mathbf{B}\mathbf{u}_n,
\label{eq:parallel}
\end{equation}
where \(\mathbf{P}\boldsymbol{\Lambda}\mathbf{P}^{-1}\) denotes the eigendecomposition. 
Unrolling this recursion yields the explicit solution
\begin{equation}
\mathbf{x}_n 
= \sum_{k=0}^{n-1} 
\mathbf{A}^k \mathbf{B}\mathbf{u}_{n-k}
= \sum_{k=0}^{n-1} 
\mathbf{P}\boldsymbol{\Lambda}^k\mathbf{P}^{-1}\mathbf{B}\mathbf{u}_{n-k}.
\label{eq:rolling}
\end{equation}
This standard eigenspace expansion of linear recurrences has also been used in oscillatory 
state-space models, such as LinOSS \citep{rusch2025oscillatory}, which focus on second-order 
ODE dynamics. Here, we adopt a similar formulation but extend it to PDE-driven spatio-temporal 
dynamics in BioOSS.

To ensure stability, all eigenvalues must satisfy \(|\lambda| \leq 1\). We begin by observing that the real eigenvalue  
\(
\lambda_1 = \frac{1}{1+\Delta t k^o_{i,j}} < 1
\)  
is strictly stable for any positive damping coefficient \(k^o_{i,j} > 0\).

For the complex conjugate eigenvalue pair \(\lambda_{2,3}\), we compute their squared magnitude:
\begin{equation}
\small
|\lambda_{2,3}|^2 = \left( \frac{1}{2} \left( \frac{1}{1 + \Delta t k^p_{i,j}} + \frac{1}{1 + \Delta t k^o_{i,j}} \right) \right)^2 
+ \frac{c_{i,j}^2 \Delta t^2 (\xi_x^2 + \xi_y^2)}{(1 + \Delta t k^p_{i,j})(1 + \Delta t k^o_{i,j})}.
\end{equation}

When both damping coefficients are positive (\(k^p_{i,j} > 0\) and \(k^o_{i,j} > 0\)), the first term remains less than 1, while the second term reflects the contribution of wave propagation.

To ensure unconditional stability over all spatial frequencies \((\xi_x, \xi_y)\), the wave speed must satisfy the following constraint, which generalizes the classic CFL (Courant–Friedrichs–Lewy) condition by incorporating the effects of damping parameters into the stability criterion:
\begin{equation}
c_{i,j} \leq \frac{\Delta x}{\Delta t} \cdot \frac{\sqrt{(1 + \Delta t k^p_{i,j})(1 + \Delta t k^o_{i,j})}}{\sqrt{2}},
\end{equation}
where \( \Delta x\) denotes fixed spatial step.

\subsection{Eigenfrequency Structure and Emergent Selectivity}
\label{subsec:eigenstructure}

The eigenvalue structure of the BioOSS system provides insight into its emergent oscillatory behavior across the spatial grid. Each pair of complex conjugate eigenvalues corresponds to a spatio-temporal mode governed by the local wave parameters and spatial frequency components. While these modes are not tied to individual neurons, their frequency profiles shape the collective dynamics of the \( p \)-field. Although eigenvalues are not explicitly computed or used during training or inference, analyzing them post hoc reveals the types of oscillations the model can support after learning.

In discrete-time systems, complex eigenvalues \(\lambda = a + ib\) are typically expressed in polar form as \(\lambda = re^{i\theta}\), where \(r = \sqrt{a^2 + b^2}\) denotes the magnitude and \(\theta = \tan^{-1}(b/a)\) is the phase angle. This angle is directly related to the angular frequency of oscillation. As complex eigenvalues arise in conjugate pairs, a half revolution (\(\pi\)) around the unit circle corresponds to the system's sampling frequency \(\frac{1}{\Delta t}\), allowing us to map the eigenvalue phase to a temporal frequency as
\(
f = \frac{\theta}{\pi\Delta t}
\).

For the dominant eigenvalues \( \lambda_{2,3} \) derived from Eq.~\eqref{eq:lambda_2_3}, this yields the following closed-form expression for the local oscillation frequency at each grid point \((i,j)\):
\begin{equation}
\small
f_{i,j} = \frac{1}{\pi\Delta t} \tan^{-1}\left( 
\frac{2 c_{i,j} \Delta t \sqrt{\xi_x^2 + \xi_y^2}}{
\sqrt{(1+\Delta tk^p_{i,j})(1+\Delta t k^o_{i,j})} 
\left( \frac{1}{1+\Delta tk^p_{i,j}} + \frac{1}{1+\Delta t k^o_{i,j}} \right)}
\right).
\label{eq:eigen_map_to_freq}
\end{equation}

This frequency mapping links each spatio-temporal mode’s oscillatory profile to the local parameters \(c\), \(k^p\), \(k^o\) and the wave vector \((\xi_x, \xi_y)\). To empirically validate this interpretation, we assess whether neurons initialized with distinct eigenmodes exhibit the expected frequency selectivity.

\begin{figure}[htbp]
    \centering
        \begin{minipage}{0.24\textwidth}
        \centering
        \includegraphics[width=\textwidth, trim=80 50 80 80, clip]{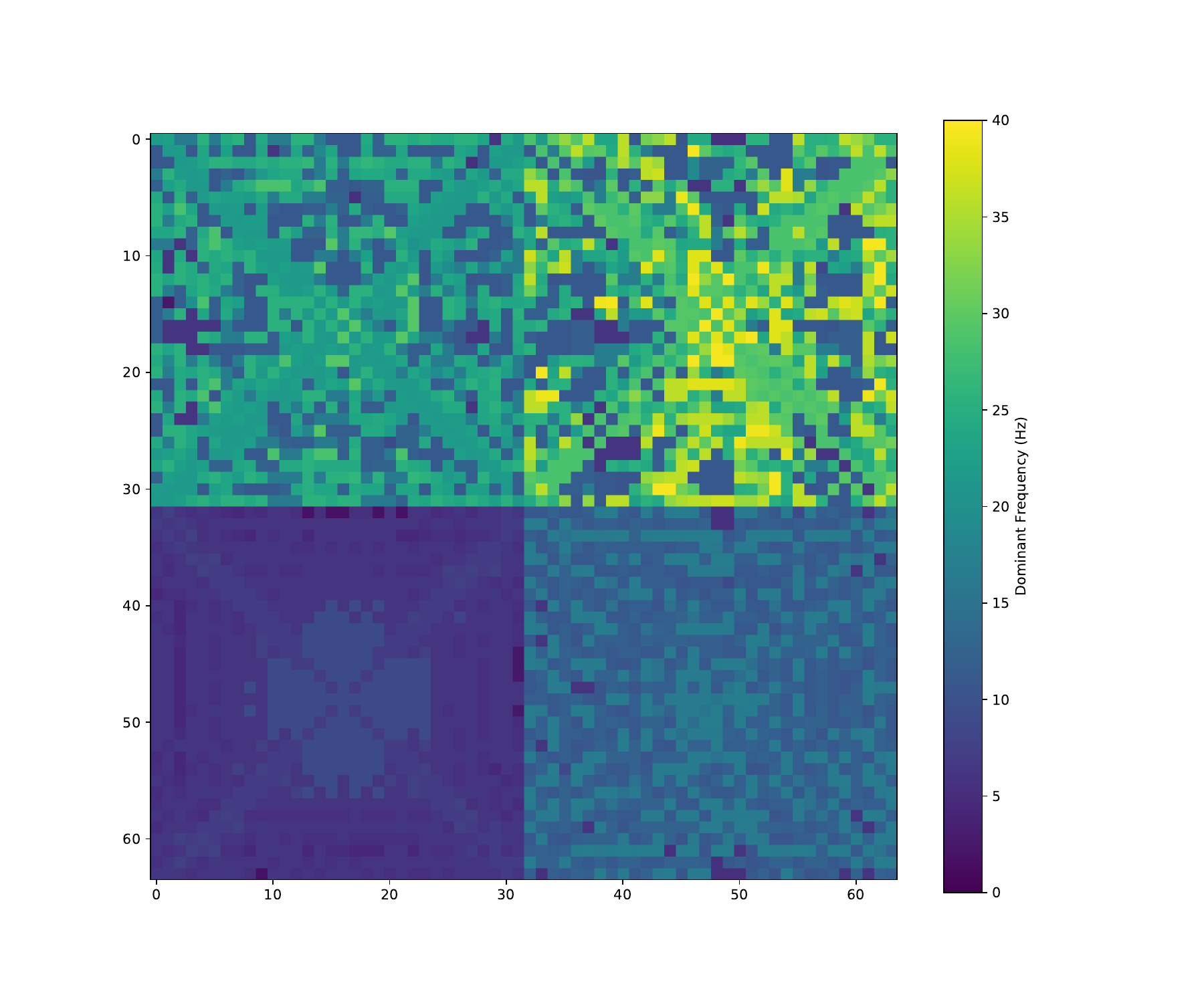}
        \subcaption{}
        \label{fig:spatial_response_a}
    \end{minipage}%
    \begin{minipage}{0.24\textwidth}
        \centering
        \includegraphics[width=\textwidth, trim=80 40 80 80, clip]{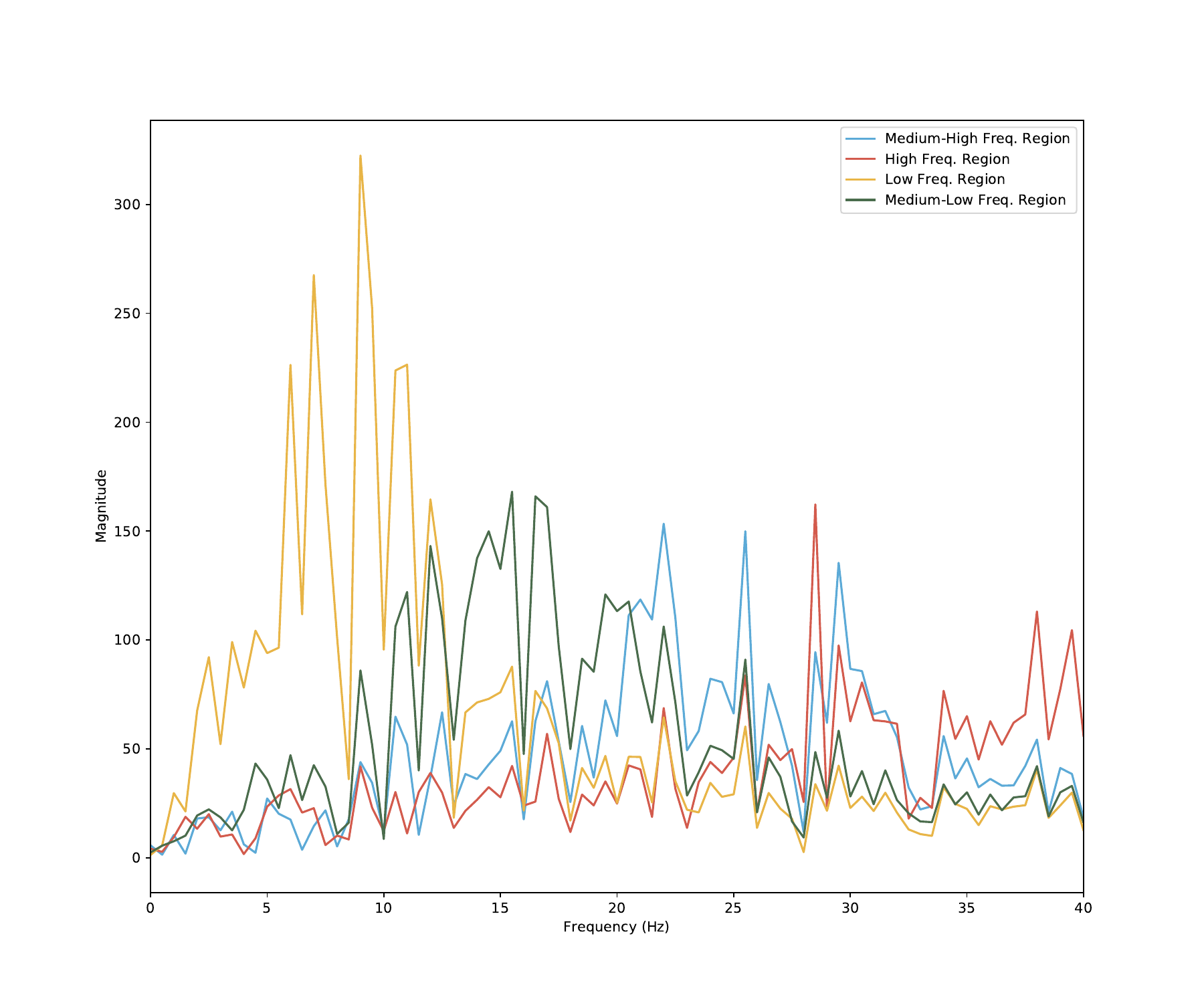}
        \subcaption{}
        \label{fig:spatial_response_b}
    \end{minipage}%
    \begin{minipage}{0.24\textwidth}
        \centering
        \includegraphics[width=\textwidth, trim=80 40 80 80, clip]{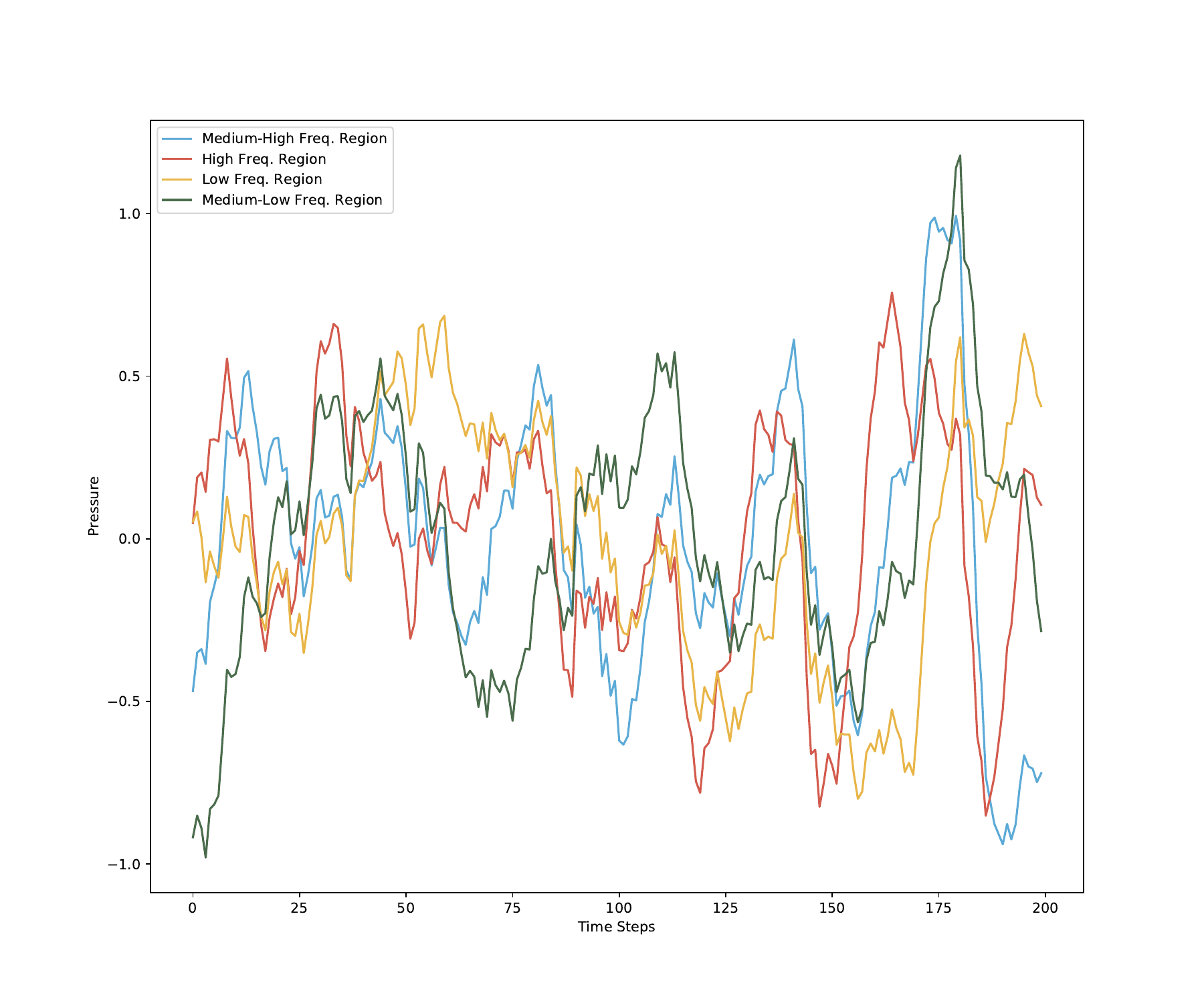}
        \subcaption{}
        \label{fig:spatial_response_c}
    \end{minipage}%
    \begin{minipage}{0.24\textwidth}
        \centering
        \includegraphics[width=\textwidth, trim=80 50 80 80, clip]{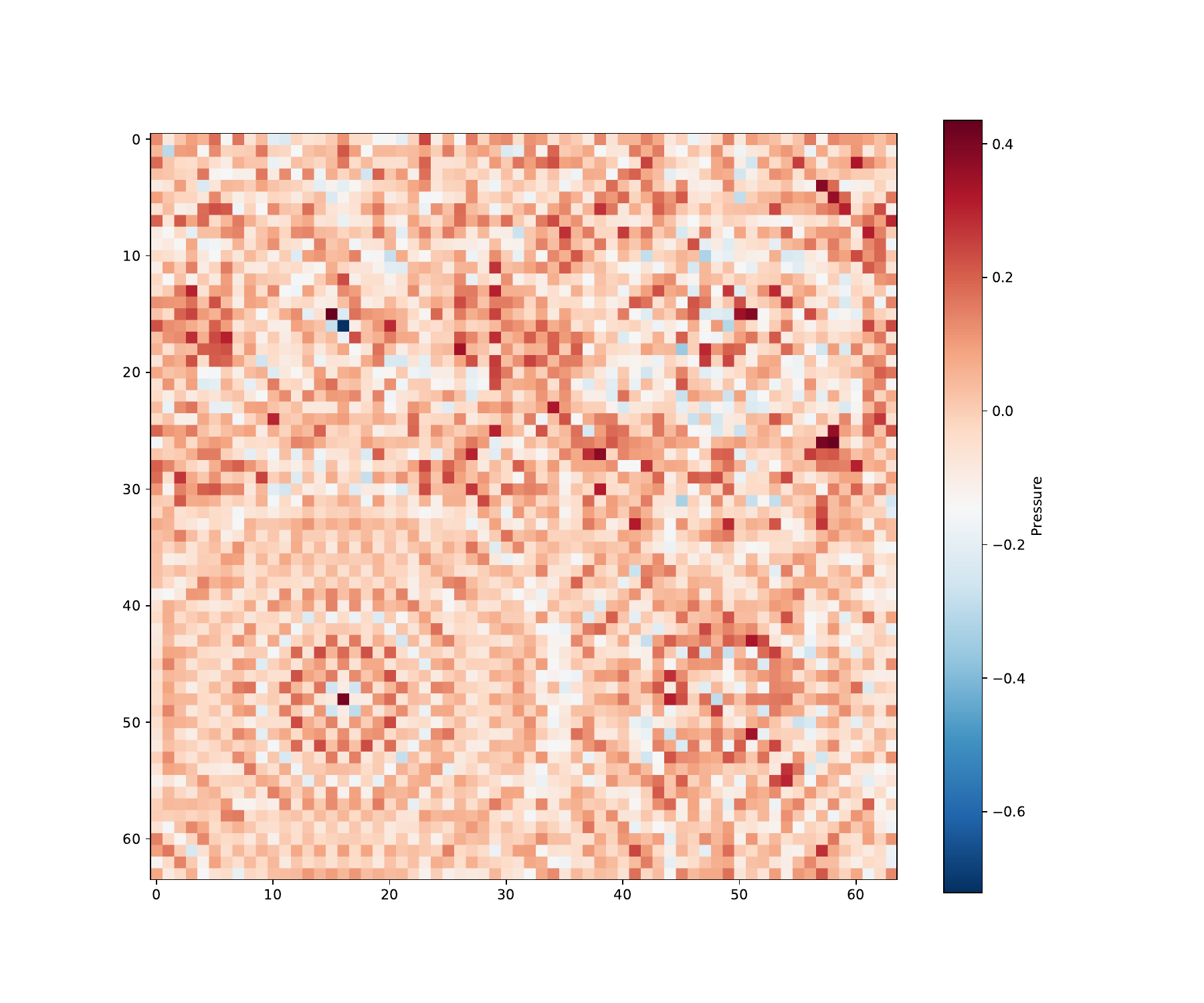}
        \subcaption{}
        \label{fig:spatial_response_d}
    \end{minipage}
    \caption{(a) Dominant-frequency map; (b) Spectra at region centers; (c) Time traces (last 200 steps); (d) Final \(p\)-field. Quadrants are initialized for 0–10, 10–20, 20–30, and 30–40~Hz and each resonates in its assigned band despite identical white-noise input, supporting the eigenvalue–frequency mapping. Full spectra and time traces are provided in Appendix~\ref{appendix:add_results}.}
    \label{fig:spatial_response_main}
\end{figure}

We partition a \(64 \times 64\) grid into four quadrants and initialize each with eigenmodes targeting a different frequency band (0–10, 10–20, 20–30, 30–40~Hz), using Eq.~\eqref{eq:eigen_map_to_freq}. While Eq.~(21) defines a mapping from eigenvalue phase to frequency using $\Delta t$, this interpretation is only meaningful when $\Delta t$ corresponds to a physical time unit. In our synthetic experiments (e.g., white-noise input with frequency-limited bands), we set $\Delta t = 10$\,ms to reflect biologically plausible temporal resolution, allowing the resulting frequencies to be interpreted in Hz. In contrast, for downstream tasks such as those in Section~\ref{section:results}, $\Delta t$ is treated as a unitless index (e.g., one step = one hour in the Traffic dataset), and frequency should be understood as a relative, dimensionless measure rather than a physical quantity. All regions receive the same white-noise input. Despite identical stimulation, each quadrant exhibits a dominant-frequency profile matching its assigned band (Fig.~\ref{fig:spatial_response_main}). This supports the hypothesis that the eigenvalue phase determines local spectral responses, and demonstrates BioOSS’s capacity for encoding frequency preferences through spatially structured initialization.

These results offer a mechanistic account of how spatio-temporal selectivity arises in BioOSS, demonstrating that distinct frequency profiles can emerge even without anatomically imposed groupings. The resulting frequency-selective regions resemble functional “islands” with coherent internal dynamics and spectral specialization. While BioOSS does not currently enforce explicit connectivity boundaries, similar modular topologies—defined by dense intra-cluster and sparse inter-cluster connections—have been shown to support self-sustained, wiring-efficient dynamics in both biological and artificial systems~\cite{meunier2010modular, liang2022less}.

\subsection{Computational Complexity}
The computational complexity of BioOSS scales linearly with the sequence length $T$, 
spatial size $HW$, and number of layers $L$. Specifically, the inference cost is 
$\mathcal{O}(LTHWd)$, and the training cost is 
$\mathcal{O}(LTHW(d + d^2))$, where $d=HW$ denotes the dimension of the hidden state. 
A detailed derivation of these results is provided in Appendix~\ref{app:complexity}.

\section{Results}

\subsection{Experimental Setting}
\paragraph{Baseline} This paper primarily focuses on the exploration of oscillatory-based spatio-temporal model and sequential tasks. Therefore, we do not include Transformer models, as they are end-to-end frameworks. BioOSS is evaluated against several representative sequential models with recurrent mechanisms, including the Linear Recurrent Unit (LRU)~\citep{orvieto2023resurrecting}, S5~\citep{smith2022simplified}, and LinOSS~\citep{rusch2025oscillatory}. While alternative sequential modeling approaches, such as the Neural Refined Differential Equation (NRDE)~\citep{irie2022neural} and the Neural Controlled Differential Equation (NCDE)~\citep{kidger2020neural}, are available, these require solving differential equations at each time step, leading to significant computational and time overhead. As such, we exclude these models from our comparison to ensure a fair and efficient benchmarking protocol.

\paragraph{Nonlinear Components} 
The core BioOSS system is linear, but practical implementations include nonlinear components such as multilayer readouts and self-gating blocks in the model head~\citep{rusch2025oscillatory}. These additions are independent of the oscillatory dynamics and are described in Appendix~\ref{appendix:training details}.

\paragraph{Hyperparameters and Environment} All BioOSS models employed in the experiments adopt a 2D architecture, with the model size varying across tasks. For each task, the learning rate was selected through a grid search to ensure optimal performance. A comprehensive summary of all hyperparameter configurations is provided in the Appendix Section~\ref{appendix:training details} Table~\ref{tab:best_config}. All experiments were conducted using JAX for classification tasks and PyTorch for prediction tasks to ensure consistency with established baselines, datasets, and evaluation protocols. Models were trained using the Adam optimizer and executed on a single NVIDIA GeForce RTX 4090 GPU (24 GB memory). Details on model parameters, GPU usage, and memory consumption are provided in Appendix~\ref{appendix:training details}, Tables~\ref{tab:exp_config_hardware_1} and~\ref{tab:exp_config_hardware_2}.

\subsection{Evaluation on Time Series Tasks}
\label{section:results}
The benchmark datasets considered in this section use time steps that do not always correspond to fixed physical durations. For example, in the Traffic dataset, each step represents one hour. As a result, while BioOSS captures oscillatory dynamics across time steps, no mapping to physical frequency units (e.g., Hz) is applied in these experiments. We evaluate the proposed BioOSS model on two fundamental time series tasks: classification and prediction. Experiments are conducted on recent benchmark datasets to demonstrate the effectiveness of the bio-inspired architecture.

All models were trained under identical settings, with results averaged over five seeds and a 
70:15:15 train/validation/test split. For BioOSS, we tuned the learning rate, while for LinOSS and 
other baselines we relied on the official implementations \citep{rusch2025oscillatory} without grid 
search. We note that our reproduced LinOSS results differ slightly from those reported in the 
LinOSS paper, which we attribute to environment dependencies, lack of hyperparameter retuning, 
and the strong sensitivity of UEA and PPG datasets to the exact random split across Python/JAX 
environments. Our aim was not to re-evaluate LinOSS, but to benchmark all models under a unified 
protocol, presenting BioOSS as a competitive and interpretable alternative rather than a new state 
of the art.

\subsubsection{Classification}
We evaluate the proposed BioOSS model on a recently introduced sequential benchmark presented 
in~\citep{walker2024log}, using the same benchmark protocol as in the LinOSS paper 
\citep{rusch2025oscillatory} to ensure fair comparison and evaluation. This benchmark includes six 
multivariate time series datasets from the University of East Anglia (UEA) Multivariate Time Series 
Classification Archive (UEA-MTSCA)~\citep{bagnall2018uea}: EigenWorms (Worms), 
SelfRegulationSCP1 (SCP1), SelfRegulationSCP2 (SCP2), EthanolConcentration (Ethanol), Heartbeat, 
and MotorImagery (Motor). These datasets exhibit a wide range of sequence lengths and classification 
complexities, providing a comprehensive evaluation of model performance under varying temporal and 
structural characteristics. The results are summarized in Table~\ref{tab:TSC_results}.

\begin{table}[htbp]
\centering
\begin{tabular}{ccccccc}
\hline
& Worms & SCP1 & SCP2 & Ethanol & Heartbeat & Motor \\
\hline
Seq. length & 17{,}984 & 896 & 1{,}152 & 1{,}751 & 405 & 3{,}000 \\
\#Classes & 5 & 2 & 2 & 4 & 2 & 2 \\
\hline
LRU & 89.4 $\pm$ 5.4 & 86.8 $\pm$ 3.1 & 54.7 $\pm$ 10.1 & 28.4 $\pm$ 5.0 & 70.0 $\pm$ 3.6 & 55.8 $\pm$ 6.2 \\
S5 & 83.9 $\pm$ 4.1 & \textcolor{red}{\textbf{86.8 $\pm$ 4.5}} & 51.9 $\pm$ 4.5 & 24.3 $\pm$ 4.8 & \textcolor{blue}{\underline{73.9 $\pm$ 3.1}} & 50.5 $\pm$ 4.2 \\
LinOSS & \textcolor{blue}{\underline{92.2 $\pm$ 7.7}} & \textcolor{blue}{\underline{85.9 $\pm$ 2.9}} & \textcolor{red}{\textbf{60.4 $\pm$ 5.7}} & \textcolor{blue}{\underline{29.9 $\pm$ 0.6}} & 71.9 $\pm$ 2.6 & \textcolor{red}{\textbf{60.4 $\pm$ 7.2}} \\
\hline
BioOSS & \textcolor{red}{\textbf{92.8 $\pm$ 5.2}} & 85.6 $\pm$ 3.9 & \textcolor{blue}{\underline{55.1 $\pm$ 1.8}} & \textcolor{red}{\textbf{33.4 $\pm$ 10.7}} & \textcolor{red}{\textbf{74.8 $\pm$ 2.0}} & \textcolor{blue}{\underline{55.8 $\pm$ 5.8}} \\
\hline
\end{tabular}
\caption{\textbf{Multivariate long-term time series classification results.} Accuracy (\%) and standard deviation for each model on six UEA datasets. Best results are shown in \textcolor{red}{\textbf{red and bold}}; second-best results are in \textcolor{blue}{\underline{blue and underlined}}. Higher is better.}
\label{tab:TSC_results}
\end{table}

We report the multivariate time series classification performance of all models across six benchmark UEA datasets in Table~\ref{tab:TSC_results}. On average, BioOSS achieves the highest accuracy at 66.25\%, outperforming all baselines, followed by LinOSS at 65.12\%. BioOSS achieves state-of-the-art results on three datasets (Worms, EthanolConcentration, and Heartbeat) and secures the second-best performance on two others (SCP2 and MotorImagery).

Notably, on the challenging Heartbeat dataset, BioOSS attains the best accuracy of 74.8\%, improving upon LinOSS and S5 by 2.9\% and 0.9\%, respectively. On the long-sequence EthanolConcentration dataset (1,751 steps), BioOSS also outperforms all baselines, achieving 33.4\% accuracy compared to 29.9\% for LinOSS and 28.4\% for LRU.

In contrast, S5 shows competitive performance only on SCP1 and Heartbeat, while LRU performs best only on SCP1 and exhibits high variance across datasets. LinOSS demonstrates consistent robustness, achieving the highest accuracy on SCP2 (60.4\%) and the second-best result on Worms (92.2\%). These results highlight the strong generalization ability of BioOSS, particularly for long sequences and heterogeneous biological and environmental signals.

\subsubsection{Prediction}
We evaluate the proposed models on the PPG-DaLiA dataset, a benchmark for heart rate regression using wearable sensor data. The dataset contains multi-channel physiological and motion signals collected from fifteen subjects engaged in daily activities. Each subject's recordings span approximately 2.5 hours at 128~Hz, covering six input channels: blood volume pulse, electrodermal activity, body temperature, and triaxial acceleration. To handle the long-range dependencies, we segment the data using a sliding window of length 49,920 and a step size of 4,992. Following the same evaluation protocol as in the LinOSS paper \citep{rusch2025oscillatory}, all models are trained with identical hyperparameter tuning procedures to ensure fair comparison and evaluation.
\begin{table}[htbp]
\begin{minipage}{\textwidth}
\centering
\begin{tabular}{lr}
\hline
Model & MSE $\times 10^{-2}$ \\
\hline
LRU & $15.64 \pm 1.67$ \\
S5 & $12.79 \pm 0.72$ \\
LinOSS & \textcolor{red}{\textbf{6.4 $\pm$ 0.23}} \\
\hline
BioOSS & \textcolor{blue}{\underline{7.7 $\pm$ 0.2}} \\
\hline
\end{tabular}
\caption{\textbf{PPG-based regression results on the PPG-DaLiA dataset.} Mean-squared error ($\times 10^{-2}$) and standard deviation over 5 training runs for each model. All models are trained using the same hyper-parameter tuning protocol to ensure fair comparison. Best results are shown in \textcolor{red}{\textbf{red and bold}} and second-best results are in \textcolor{blue}{\underline{blue and underlined}}. Lower is better.}
\label{tab:TSP_ppg_results}
\end{minipage}
\end{table}
Table~\ref{tab:TSP_ppg_results} summarizes the average test mean-squared error (MSE) over five independent runs. LinOSS achieves the lowest MSE, while BioOSS secures the second-best performance. Compared to traditional recurrent and state-space baselines such as LRU and S5, both LinOSS and BioOSS demonstrate clear advantages in modeling long-duration dependencies, highlighting the effectiveness of the proposed structured designs for physiological signals.

To further evaluate predictive capabilities, we assess the models on four additional benchmark datasets characterized by clear periodic patterns: Electricity, Solar-Energy, Traffic, and Weather~\citep{lai2018modeling}. Dataset splits, look-back sequence lengths, and prediction horizons are aligned with the cyclic properties of each dataset. The results, summarized in Table~\ref{tab:TSP_benchmark_results}, further validate the strong performance of LinOSS and BioOSS across diverse long-term forecasting tasks.

\begin{table}[htbp]
\centering
\begin{tabular}{ccccc}
\hline
& Electricity & Solar-Energy & Traffic & Weather \\
\hline
Timesteps & 26{,}304 & 52{,}560 & 17{,}544 & 52{,}696 \\
Channels & 321 & 137 & 862 & 21 \\
Frequency & 1 hour & 10 mins & 1 hour & 10 mins \\
Cyclic Patterns & Daily \& Weekly & Daily & Daily \& Weekly & Daily \\
Cycle Length & 168 & 144 & 168 & 144 \\
\hline
LinOSS & 21.875 $\pm$ 0.000 & 22.363 $\pm$ 0.000 & 68.828 $\pm$ 0.002 & 22.288 $\pm$ 0.000 \\
BioOSS & \textcolor{red}{\textbf{18.699 $\pm$ 0.000}} & \textcolor{red}{\textbf{21.873 $\pm$ 0.003}} & \textcolor{red}{\textbf{62.284 $\pm$ 0.007}} & \textcolor{red}{\textbf{22.232 $\pm$ 0.000}} \\
\hline
\end{tabular}
\caption{\textbf{Multivariate time-series forecasting results across four benchmark datasets with intrinsic patterns.} Mean-squared error (MSE $\times 10^{-2}$) and standard deviation over 5 training runs are reported for each model. All models are trained using the same hyper-parameter tuning protocol to ensure a fair comparison. Best results for each dataset are highlighted in \textcolor{red}{\textbf{red and bold}}. Lower is better.}
\label{tab:TSP_benchmark_results}
\end{table}

Table~\ref{tab:TSP_benchmark_results} reports the average test mean-squared error (MSE) across five independent runs for LinOSS and BioOSS on four benchmark multivariate time series datasets with inherent temporal patterns. BioOSS consistently outperforms LinOSS, achieving the lowest MSEs of 18.699, 21.873, 62.284, and 22.232 on the Electricity, Solar-Energy, Traffic, and Weather datasets, respectively. In contrast, LinOSS yields MSEs of 21.875, 22.363, 68.828, and 22.288 on the same datasets. These results highlight the superior ability of BioOSS to capture temporal dependencies, particularly in datasets exhibiting daily and weekly cyclic patterns. The oscillatory structure of BioOSS enables more effective modeling of periodic dynamics, even with a comparable number of neurons to LinOSS.

\subsection{Wave Dynamics Visualization}
To qualitatively examine the internal dynamics of BioOSS beyond benchmark accuracy, we generated visualizations of emergent wave behavior in the 2D oscillatory layer. Specifically, normalized logits were extracted from an audio tagging backbone processing a 1-minute audio segment using a 4-second sliding window with a 1-second stride. These logits were then used as point-wise inputs to BioOSS, with each logit dimension mapped to a local patch of 7--8 grid cells. The system subsequently evolved under BioOSS dynamics, and the pressure-field ($p$) values were recorded over time. The resulting sequences reveal traveling-wave propagation and coupling effects: activity initiated in one region not only spreads across its assigned area but also induces oscillatory responses in neighboring regions through local interactions. This dynamic behavior complements the static examples shown in Fig.~\ref{fig:overview}(d--e), providing direct evidence of emergent spatiotemporal coordination. Animated GIFs of representative sequences are provided in the Supplementary Material, which contains three audio samples illustrating the evolution of the $p$-field under realistic inputs.

\section{Conclusion}
We proposed BioOSS, a biologically inspired oscillatory state system that captures spatio-temporal dynamics through wave-like propagation between pressure and velocity-like neurons on a 2D grid. By introducing trainable damping and propagation parameters, BioOSS adapts to task-specific patterns while preserving numerical stability and interpretability. It achieves strong performance across classification and forecasting benchmarks, particularly excelling in frequency-selective and long-range temporal tasks. Unlike traditional models, BioOSS offers insight into its internal dynamics via eigenvalue-based frequency decomposition. While promising, future work can explore scaling, multimodal integration, and continuous learning. BioOSS offers a step toward interpretable, brain-inspired computation in AI. A detailed discussion of limitations is included in Appendix~\ref{appendix:limit}.

\section*{Acknowledgment}
This work was supported by the Special Research Fund (BOF) of Ghent University under Grant BOF/24J/2021/246, and by the Flemish Government through the Onderzoeksprogramma AI Vlaanderen programme.


\bibliographystyle{plainnat}
\bibliography{reference}

\newpage

\appendix

\section{Fourier Transformation of Differential Operators}
\label{appendix:Fourier transform}

In this section, we derive the fundamental relationship between differential operators in physical space and their counterparts in Fourier space.

\subsection{Definitions and Notation}


We begin with the definition of the Fourier transform for a function \( p(\mathbf{r}) \) in \( \mathbb{R}^n \):
\begin{equation}
    \tilde{p}(\boldsymbol{\xi}) = \mathcal{F}\{p(\mathbf{r})\} = \int_{\mathbb{R}^n} p(\mathbf{r}) e^{-i\boldsymbol{\xi} \cdot \mathbf{r}} \, d\mathbf{r},
\end{equation}
where \( \boldsymbol{\xi} = [\xi_x, \xi_y, \ldots]^\top \) is the frequency domain vector.

For simplicity, we focus on the two-dimensional case where \( \mathbf{r} = [x, y]^\top \) and \( \boldsymbol{\xi} = [\xi_x, \xi_y]^\top \).

\subsection{Transformation of the Gradient Operator}

We aim to derive the Fourier transform of the gradient of \( p \), denoted by \( \nabla p \):
\begin{equation}
    \mathcal{F}\{\nabla p\} = \mathcal{F}\left\{\begin{bmatrix} \frac{\partial p}{\partial x} \\ \frac{\partial p}{\partial y} \end{bmatrix}\right\}.
\end{equation}

We first evaluate the Fourier transform of the spatial derivative of \(p\), given by \(\mathcal{F}\left\{\frac{\partial p}{\partial x}\right\}\):
\begin{equation}
    \mathcal{F}\left\{\frac{\partial p}{\partial x}\right\} = \int_{\mathbb{R}^2} \frac{\partial p(x,y)}{\partial x} e^{-i(\xi_x x + \xi_y y)} \, dx \, dy.
\end{equation}

Focusing on the integration with respect to \(x\), we apply integration by parts:
\begin{equation}
\begin{aligned}
    \int_{-\infty}^{\infty} \frac{\partial p(x,y)}{\partial x} e^{-i\xi_x x} \, dx &= \left[ p(x,y) e^{-i\xi_x x} \right]_{-\infty}^{\infty} - \int_{-\infty}^{\infty} p(x,y) \frac{\partial}{\partial x} \left( e^{-i\xi_x x} \right) \, dx \\
    &= \left[ p(x,y) e^{-i\xi_x x} \right]_{-\infty}^{\infty} - \int_{-\infty}^{\infty} p(x,y) \left( -i\xi_x e^{-i\xi_x x} \right) \, dx \\
    &= \left[ p(x,y) e^{-i\xi_x x} \right]_{-\infty}^{\infty} + i\xi_x \int_{-\infty}^{\infty} p(x,y) e^{-i\xi_x x} \, dx.
\end{aligned}
\end{equation}

For functions that are sufficiently smooth and decay rapidly as \(|x| \to \infty\), the boundary term vanishes:
\begin{equation}
    \left[ p(x,y) e^{-i\xi_x x} \right]_{-\infty}^{\infty} = 0.
\end{equation}

Therefore:
\begin{equation}
    \int_{-\infty}^{\infty} \frac{\partial p(x,y)}{\partial x} e^{-i\xi_x x} \, dx = i\xi_x \int_{-\infty}^{\infty} p(x,y) e^{-i\xi_x x} \, dx.
\end{equation}

Reintroducing the integration with respect to \( y \):
\begin{equation}
\begin{aligned}
    \mathcal{F}\left\{\frac{\partial p}{\partial x}\right\} &= \int_{-\infty}^{\infty} \int_{-\infty}^{\infty} \frac{\partial p(x,y)}{\partial x} e^{-i(\xi_x x + \xi_y y)} \, dx \, dy \\
    &= \int_{-\infty}^{\infty} \left[ i\xi_x \int_{-\infty}^{\infty} p(x,y) e^{-i\xi_x x} \, dx \right] e^{-i\xi_y y} \, dy \\
    &= i\xi_x \int_{-\infty}^{\infty} \int_{-\infty}^{\infty} p(x,y) e^{-i(\xi_x x + \xi_y y)} \, dx \, dy \\
    &= i\xi_x \tilde{p}(\xi_x, \xi_y).
\end{aligned}
\end{equation}

By symmetry, we can similarly derive:
\begin{equation}
    \mathcal{F}\left\{\frac{\partial p}{\partial y}\right\} = i\xi_y \tilde{p}(\xi_x, \xi_y).
\end{equation}

Thus, for the gradient operator, we have:
\begin{equation}
    \mathcal{F}\{\nabla p\} = \mathcal{F}\left\{\begin{bmatrix} \frac{\partial p}{\partial x} \\ \frac{\partial p}{\partial y} \end{bmatrix}\right\} = \begin{bmatrix} i\xi_x \tilde{p} \\ i\xi_y \tilde{p} \end{bmatrix} = i\boldsymbol{\xi} \tilde{p}.
\end{equation}

This gives the important relationship:
\begin{equation}
    \nabla p \xrightarrow{\mathcal{F}} i\boldsymbol{\xi} \tilde{p}.
\end{equation}

\subsection{Transformation of the Divergence Operator}

For a vector field \(\mathbf{o} = [o_x, o_y]^\top\), the divergence operator is defined as follows:
\begin{equation}
    \nabla \cdot \mathbf{o} = \frac{\partial o_x}{\partial x} + \frac{\partial o_y}{\partial y}.
\end{equation}

Applying the Fourier transform and using the result from the previous section:
\begin{equation}
\begin{aligned}
    \mathcal{F}\{\nabla \cdot \mathbf{o}\} &= \mathcal{F}\left\{\frac{\partial o_x}{\partial x} + \frac{\partial o_y}{\partial y}\right\} \\
    &= \mathcal{F}\left\{\frac{\partial o_x}{\partial x}\right\} + \mathcal{F}\left\{\frac{\partial o_y}{\partial y}\right\} \\
    &= i\xi_x \tilde{o}_x + i\xi_y \tilde{o}_y \\
    &= i\boldsymbol{\xi} \cdot \tilde{\mathbf{o}}.
\end{aligned}
\end{equation}

This gives the important relationship:
\begin{equation}
    \nabla \cdot \mathbf{o} \xrightarrow{\mathcal{F}} i\boldsymbol{\xi} \cdot \tilde{\mathbf{o}}.
\end{equation}

Thus, the Fourier transform converts differential operators into algebraic operations:
\begin{align}
    \frac{\partial}{\partial t} \xrightarrow{\mathcal{F}} -i\omega, \\
    \nabla p \xrightarrow{\mathcal{F}} i\boldsymbol{\xi} \tilde{p}, \\
    \nabla \cdot \mathbf{o} \xrightarrow{\mathcal{F}} i\boldsymbol{\xi} \cdot \tilde{\mathbf{o}}.
\end{align}




\section{Analytical Form of Eigenvectors \(\mathbf{P}\)}
\label{appendix:eigenvector}

We consider the linearized local system matrix at a grid point \((i,j)\) in Fourier space:
\[
A =
\begin{bmatrix}
\alpha & -\alpha c^2 \Delta t\, i\xi_x & -\alpha c^2 \Delta t\, i\xi_y \\
-\beta \Delta t\, i\xi_x & \beta & 0 \\
-\beta \Delta t\, i\xi_y & 0 & \beta
\end{bmatrix},
\quad
\alpha = \frac{1}{1 + \Delta t\,k^p}, \quad \beta = \frac{1}{1 + \Delta t\,k^o}.
\]

Let \(\lambda\) be an eigenvalue of \(A\), and \(\mathbf{v} = [v_1, v_2, v_3]^T\) its associated eigenvector. We assume without loss of generality \(v_1 = 1\) and solve \((A - \lambda I)\mathbf{v} = 0\). From the second and third rows, we obtain
\[
v_2 = \frac{\beta \Delta t\, i\xi_x}{\lambda - \beta}, \quad v_3 = \frac{\beta \Delta t\, i\xi_y}{\lambda - \beta}.
\]

For eigenvalues \(\lambda_2\) and \(\lambda_3\) corresponding to oscillatory modes, the associated eigenvectors are
\[
\mathbf{v}_2 =
\begin{bmatrix}
1 \\
\displaystyle \frac{\Delta t \cdot i \xi_x}{(1 + \Delta t\,k^o)(\lambda_2 - \lambda_1)} \\
\displaystyle \frac{\Delta t \cdot i \xi_y}{(1 + \Delta t\,k^o)(\lambda_2 - \lambda_1)}
\end{bmatrix},
\quad
\mathbf{v}_3 =
\begin{bmatrix}
1 \\
\displaystyle \frac{-\Delta t \cdot i \xi_x}{(1 + \Delta t\,k^o)(\lambda_3 - \lambda_1)} \\
\displaystyle \frac{-\Delta t \cdot i \xi_y}{(1 + \Delta t\,k^o)(\lambda_3 - \lambda_1)}
\end{bmatrix}.
\]
In contrast, for the purely dissipative eigenvalue \(\lambda_1 = \beta\), the eigenvectors correspond to vectors in the nullspace of the first row. A basis can be chosen, for example, as
\[
\mathbf{v}_1 =
\begin{bmatrix}
0 \\ 1 \\ 0
\end{bmatrix},
\quad \text{or} \quad
\begin{bmatrix}
0 \\ 0 \\ 1
\end{bmatrix}.
\]

\section{Computational Complexity Analysis}
\label{app:complexity}

The computational cost of BioOSS primarily arises from its spatio-temporal update 
in the hidden layer, which is based on finite-difference approximations of a coupled PDE system. 
We characterize training and inference complexity as follows.

Let $p \in \mathbb{R}^{H \times W}$ and $o=(o^x,o^y)\in \mathbb{R}^{2 \times H \times W}$ 
denote the pressure- and velocity-like states defined on a 2D grid. At each time step, 
a BioOSS layer performs:

\begin{itemize}
    \item Gradient update: 
    \[
    \nabla p_{i,j} = 
    \left(\frac{p_{i,j}-p_{i-1,j}}{\Delta x},\,
          \frac{p_{i,j}-p_{i,j-1}}{\Delta x}\right)
    \]
    \item Divergence update:
    \[
    \nabla \cdot o_{i,j} = 
    \frac{o^x_{i+1,j}-o^x_{i,j}}{\Delta x} +
    \frac{o^y_{i,j+1}-o^y_{i,j}}{\Delta x}
    \]
    \item Damped update:
    \[
    o^{(l)} = \frac{o^*}{1+\Delta t\cdot k^o}, 
    \quad p^{(l)} = \frac{p^*}{1+\Delta t\cdot k^p}
    \]
\end{itemize}

These operations require $\mathcal{O}(HW)$ computations per layer per time step, 
where $H \times W$ is the spatial resolution of the 2D state. Each layer also performs 
linear projections, gating, and readout on a hidden state of dimension $d=HW$, 
yielding a cost of $\mathcal{O}(HW \cdot d)$.

Thus, the total inference cost per layer over $T$ time steps is:
\[
\mathcal{O}(LTHWd).
\]

During training, gradients must also be computed through the differentiable spatial operators 
and projection layers, with weight updates costing $\mathcal{O}(HW \cdot d^2)$ per layer. 
Therefore, the total training complexity is:
\[
\mathcal{O}(LTHW(d+d^2)).
\]

In practice, the hidden layer dominates both inference and training cost, and the model’s 
complexity scales linearly with sequence length $T$, number of layers $L$, and spatial size $HW$.

\section{Training Details}
\label{appendix:training details}

Algorithm~\ref{alg:biooss} illustrates the full architecture of the multi-layer BioOSS model. While the main text focuses on a single-layer formulation, the full model stacks \(L\) such layers sequentially. Each layer integrates two key components: a biologically inspired spatio-temporal propagation mechanism and a scan-compatible gated update.

Given an input sequence \( u \), the hidden state \( x^{(l)} \) is first updated by solving a PDE-based wave propagation equation over a 2D neural grid, as defined in Eq.~\eqref{eq:hidden_state_PDE}. This involves local interactions between pressure-like \( p \) neurons and velocity-like \( o \) neurons through spatial gradients and divergence operators. These updates are governed by trainable wave speed \( c \) and damping coefficients \( k^p, k^o \).

The resulting hidden state is then projected through a linear layer and passed to a gating mechanism, inspired by LinOSS~\cite{rusch2025oscillatory}, but applied on a 2D spatio-temporal field. The gating update follows:
\[
x_{\text{out}}^{(l)} = \sigma(W_g x^{(l)}) \odot \tanh(W_z x^{(l)}) + (1 - \sigma(W_g x^{(l)})) \odot x^{(l-1)},
\]
where \( W_g \) and \( W_z \) are learnable matrices shared across time. This residual formulation promotes stable deep learning and supports efficient scan operations.

Next, the gated output is mapped via a linear readout: \( y^{(l)} = C x_{\text{out}}^{(l)} + D u^{(l-1)} \), followed by a GELU activation~\cite{hendrycks2016gaussian}. A Gated Linear Unit (GLU)~\cite{dauphin2017language} then modulates the activations via \(\text{GLU}(x) = \sigma(W_1 x) \circ (W_2 x)\). The result is added to the previous layer input to produce \( u^{(l)} \), which is fed into the next block.

After \( L \) layers, the final output is computed as \( o = W_{\mathrm{out}} y^{(L)} + b_{\mathrm{out}} \). All operations are parallelized across time steps, and time indices are omitted from the notation for clarity.


\begin{algorithm}
\caption{Training Procedure for Full BioOSS Model with Spatio-Temporal Layer}
\label{alg:biooss}
\begin{algorithmic}[1]
\STATE \textbf{Input:} Input sequence \(u\)
\STATE \textbf{Output:} BioOSS output sequence \(o\) after \(L\) layers
\FOR{\(l = 1\) \textbf{to} \(L\)}
    \STATE \textit{// Step 1: Spatio-temporal PDE-based propagation over 2D grid}
    \STATE Compute gradient: \(\nabla p^{(l-1)} \gets \text{SpatialGradient}(p^{(l-1)})\)
    \STATE Compute divergence: \(\nabla \cdot o^{(l-1)} \gets \text{SpatialDivergence}(o^{(l-1)})\)
    \STATE Update velocity-like neurons: \(o^\ast \gets o^{(l-1)} - \Delta t \cdot \nabla p^{(l-1)}\)
    \STATE Update pressure-like neurons: \(p^\ast \gets p^{(l-1)} - \Delta t \cdot (c^2 \odot \nabla \cdot o^{(l-1)}) + \Delta t \cdot B u^{(l-1)}\)
    \STATE Apply damping: \(o^{(l)} \gets o^\ast / (1 + \Delta t \cdot k^o), \quad p^{(l)} \gets p^\ast / (1 + \Delta t \cdot k^p)\)
    \STATE Hidden state: \(x^{(l)} \gets \text{concat}(p^{(l)}, o^{(l)})\)
    
    \STATE \textit{// Step 2: Linear projection and gated residual update (SSM-style)}
    \STATE Linear projection: \(z^{(l)} \gets W_z x^{(l)}\)
    \STATE Gating: \(\text{gate}^{(l)} \gets \sigma(W_g x^{(l)})\)
    \STATE Residual gated update: \(x_{\text{out}}^{(l)} \gets \text{gate}^{(l)} \odot \tanh(z^{(l)}) + (1 - \text{gate}^{(l)}) \odot x^{(l-1)}\)

    \STATE \textit{// Step 3: Readout and layer update}
    \STATE Linear readout: \(y^{(l)} \gets C x_{\text{out}}^{(l)} + D u^{(l-1)}\)
    \STATE Apply nonlinearity: \(y^{(l)} \gets \text{GELU}(y^{(l)})\)
    \STATE Update input for next layer: \(u^{(l)} \gets \text{GLU}(y^{(l)}) + u^{(l-1)}\)
\ENDFOR
\STATE \textbf{Return} final output: \( \mathbf{o} \leftarrow \mathbf{W}_{\mathrm{out}} \mathbf{y}^{(L)} + \mathbf{b}_{\mathrm{out}} \)
\end{algorithmic}
\end{algorithm}

The best-performing hyperparameter configurations for the BioOSS model across all datasets are summarized in Table~\ref{tab:best_config}. These configurations were obtained using the same search procedure described in the main paper (Section~\ref{section:results}). Due to the high nonlinearity of the BioOSS model, its performance is highly sensitive to both the learning rate and the spatial grid resolution. Therefore, for each dataset, we conducted a systematic grid search, exploring learning rates from 0.000001 to 0.005, hidden dimensions of 16, 32, 64, 128, and 256, and spatial grid sizes of 10\(\times\)10, 15\(\times\)15, and 20\(\times\)20. For each dataset, we report the selected learning rate, hidden dimension size, spatial grid resolution, and number of blocks. Notably, a single block was sufficient for all cases, while the optimal learning rate, hidden size, and grid size varied depending on the dataset characteristics.

\begin{table}[ht]
\centering
\begin{tabular}{ccccc}
\hline
\textbf{Model} & \textbf{lr} & \textbf{Hidden Dim} & \textbf{Grid Size} & \textbf{\#layers} \\
\hline
Worms & 0.0001 & 128 & 20 \( \times \) 20 & 1 \\
SCP1 & 0.0002 & 128 & 20 \( \times \) 20 & 1 \\
SCP2 & 0.00001 & 128 & 15 \( \times \) 15 & 1 \\
Ethanol & 0.00025 & 16 & 20 \( \times \) 20 & 1 \\
Heartbeat & 0.000005 & 16 & 20 \( \times \) 20 & 1 \\
Motor & 0.000004 & 128 & 20 \( \times \) 20 & 1 \\
PPG & 0.0002 & 128 & 20 \( \times \) 20 & 1 \\
Electricity & 0.01 & 16 & 10 \( \times \) 10 & 1 \\
Solar-Energy & 0.01 & 16 & 10 \( \times \) 10 & 1 \\
Traffic & 0.01 & 16 & 10 \( \times \) 10 & 1 \\
Weather & 0.01 & 16 & 10 \( \times \) 10 & 1 \\
\hline
\end{tabular}
\caption{Best hyperparameter configurations for BioOSS.}
\label{tab:best_config}
\end{table}

We report the number of parameters, GPU memory usage (in MB), and run time (in seconds) for all evaluated models across the datasets introduced in Section~\ref{section:results}. The experimental settings for the baseline models strictly follow those established in~\citep{rusch2025oscillatory}, ensuring consistency in both codebase and hardware environment. All experiments were conducted on the same GPU device to guarantee fair and direct comparability. Table~\ref{tab:exp_config_hardware_1} summarizes the model configurations, memory consumption, and training run time for the time-series classification (TSC) and PPG forecasting tasks, where all experiments were implemented using JAX. In contrast, Table~\ref{tab:exp_config_hardware_2} presents the corresponding results for the long-term time-series forecasting (TSP) benchmarks, with experiments conducted in PyTorch. We note that differences in parallel computation mechanisms between JAX and PyTorch may account for minor variations in GPU memory usage. Overall, LinOSS and BioOSS exhibit competitive memory efficiency and favorable run time performance relative to established baselines such as LRU and S5.

\begin{table}[htbp]
\centering
\small
\begin{tabular}{llrrrr}
\toprule
\textbf{Dataset} & \textbf{Config} & \textbf{LRU} & \textbf{S5} & \textbf{LinOSS} & \textbf{BioOSS} \\
\midrule
\multirow{3}{*}{Worms}
  & \#parameters     & 101129 & 22007  & 134279 & 153549
 \\
  & GPU memory (MB)  & 18761  & 18877   & 18747  & 18727 \\
  & run time (s)     & 138     & 78     & 11     & 15\\
\midrule
\multirow{3}{*}{SCP1}
  & \#parameters     & 25892  & 226328 & 991240 & 153162
 \\
  & GPU memory (MB)  & 18759  & 18879  & 18893   & 18739 \\
  & run time (s)     & 15     & 4      & 34    & 6 \\
\midrule
\multirow{3}{*}{SCP2}
  & \#parameters     & 26020  & 5652   & 448072 & 107002 \\
  & GPU memory (MB)  & 18781  & 18881   & 18853   &  18741 \\
  & run time (s)     & 10     & 3      &  46    & 4 \\
\midrule
\multirow{3}{*}{Ethanol}
  & \#parameters     & 76522  & 76214  & 6728   & 16332
 \\
  & GPU memory (MB)  & 18737  & 18915   & 18813   & 18733 \\
  & run time (s)     & 4      & 5     &  4     & 9 \\
\midrule
\multirow{3}{*}{Heartbeat}
  & \#parameters     & 338820 & 158310 & 10936  & 16300 \\
  & GPU memory (MB)  & 18503   & 18885   & 18835  & 18741 \\
  & run time (s)     & 7     & 3     &  5     & 2 \\
\midrule
\multirow{3}{*}{Motor}
  & \#parameters     & 107544 & 17496  & 91844  & 153162 \\
  & GPU memory (MB)  & 18753  & 18857   & 18769   & 18727 \\
  & run time (s)     & 47    &  26    &  8    & 17 \\
\midrule
\multirow{3}{*}{PPG}
  & \#parameters     & 107544 & 17496  & 91844  & 169416 \\
  & GPU memory (MB)  & 18773   & 18915  & 18719   & 18729 \\
  & run time (s)     & 107     & 57    & 21     & 17 \\
\bottomrule
\end{tabular}
\caption{Number of parameters, GPU memory usage (in MB) and run time (in seconds) for selected models on all datasets from Table~\ref{tab:TSC_results} and Table~\ref{tab:TSP_ppg_results} in Section~\ref{section:results}.}
\label{tab:exp_config_hardware_1}
\end{table}

\begin{table}[htbp]
\centering
\small
\begin{tabular}{llrrrr}
\toprule
\textbf{Model} & \textbf{Config} & \textbf{Electricity} & \textbf{Solar-Energy} & \textbf{Traffic} & \textbf{Weather} \\
\midrule
\multirow{3}{*}{LinOSS}
  & \#parameters     & 91521  &   67785    &  161310  & 52821 \\
  & GPU memory (MB)  & 1871  & 1717   & 2079   & 1725  \\
  & run time (s)     & 89    & 70     & 105     &  71    \\
\midrule
\multirow{3}{*}{BioOSS}
  & \#parameters     & 101129 & 22007  & 26119  & 134279 \\
  & GPU memory (MB)  & 1765   & 1659   & 1967   & 1653  \\
  & run time (s)     & 170    & 170     & 196   & 162     \\
\bottomrule
\end{tabular}
\caption{Number of parameters, GPU memory usage (in MB) and run time (in seconds) for selected models on all datasets from Table~\ref{tab:TSP_benchmark_results} in Section~\ref{section:results}.}
\label{tab:exp_config_hardware_2}
\end{table}


\section{Additional Spatio-Temporal Dynamics Related Results}
\label{appendix:add_results}

\begin{figure}[htbp]
    \centering
    \begin{minipage}{0.24\textwidth}
        \centering
        \includegraphics[width=\textwidth, trim=80 50 80 80, clip]{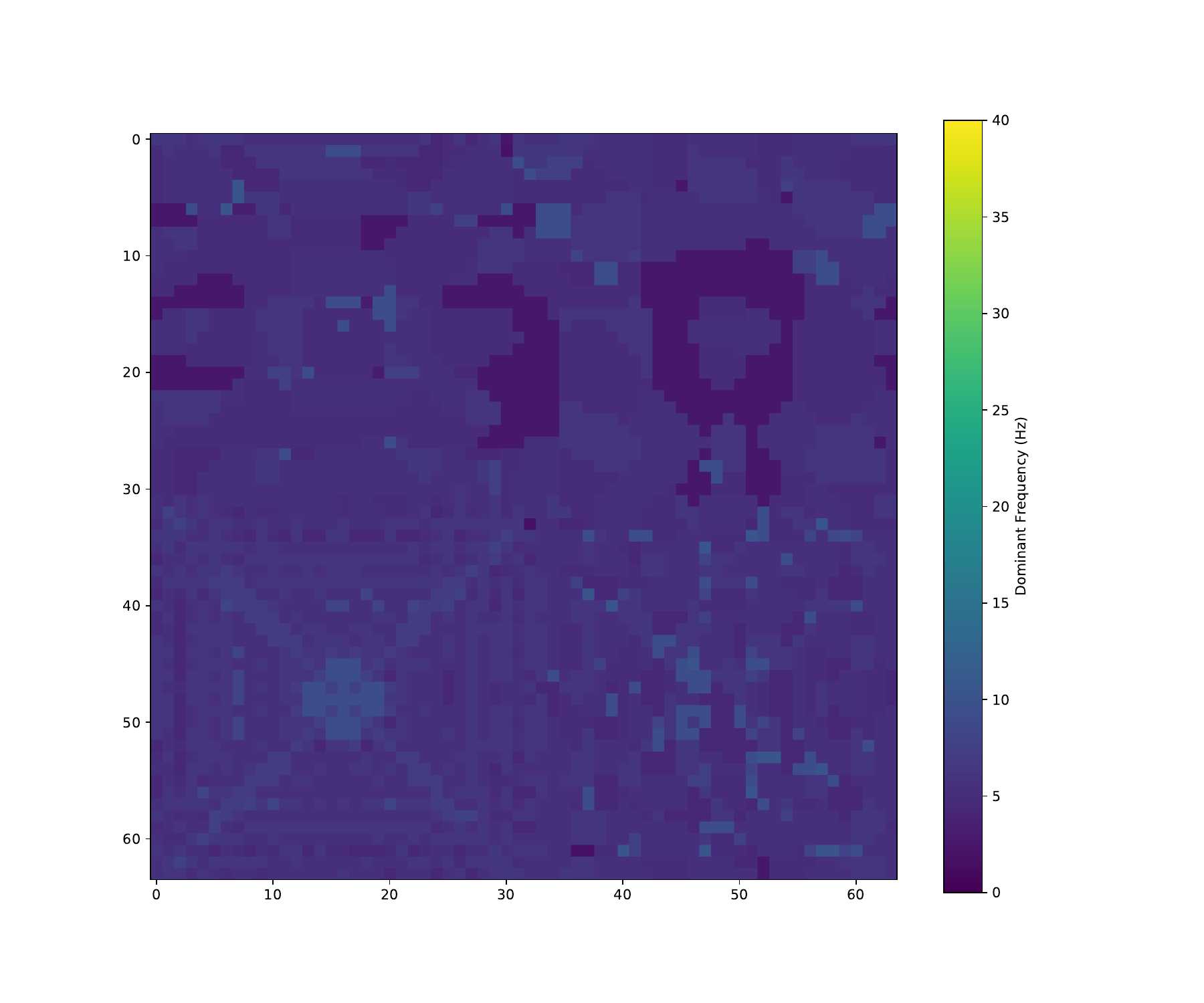}
        \subcaption{}
        \label{fig:spatial_response_a_app}
    \end{minipage}%
    \begin{minipage}{0.24\textwidth}
        \centering
        \includegraphics[width=\textwidth, trim=80 40 80 80, clip]{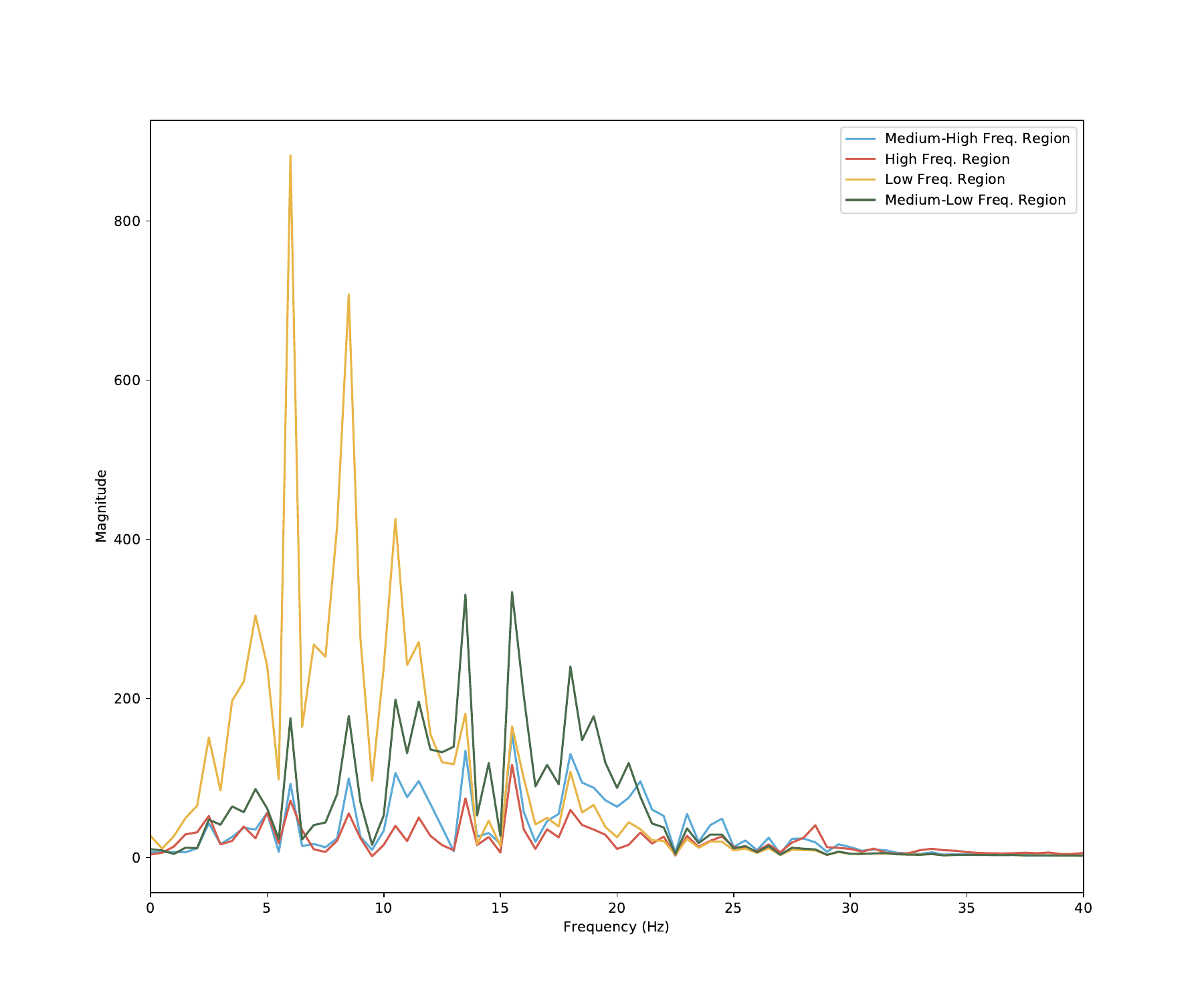}
        \subcaption{}
        \label{fig:spatial_response_b_app}
    \end{minipage}%
    \begin{minipage}{0.24\textwidth}
        \centering
        \includegraphics[width=\textwidth, trim=80 40 80 80, clip]{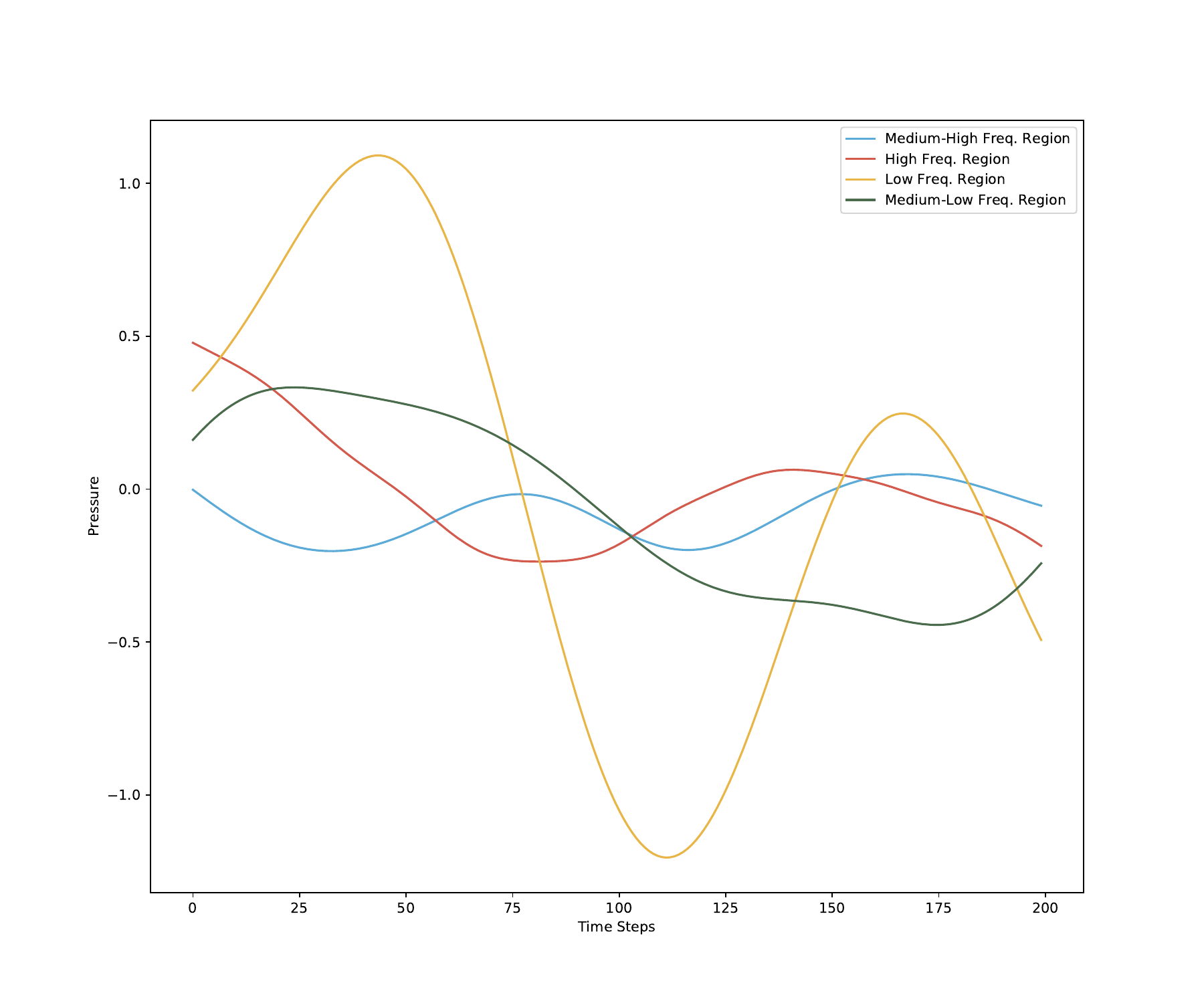}
        \subcaption{}
        \label{fig:spatial_response_c_app}
    \end{minipage}%
    \begin{minipage}{0.24\textwidth}
        \centering
        \includegraphics[width=\textwidth, trim=80 50 80 80, clip]{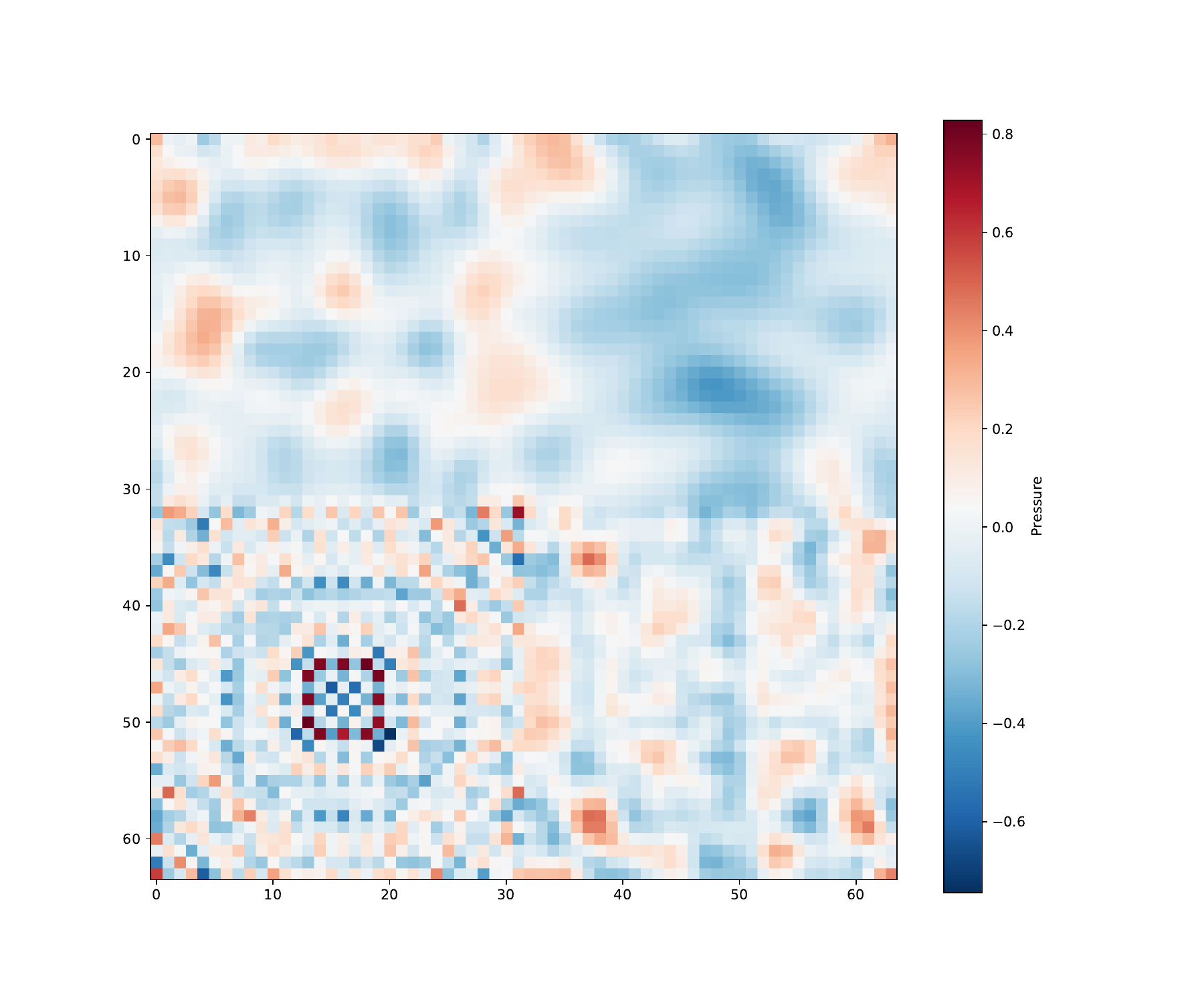}
        \subcaption{}
        \label{fig:spatial_response_d_app}
    \end{minipage}

    \vspace{0.5em} 

    \begin{minipage}{0.24\textwidth}
        \centering
        \includegraphics[width=\textwidth, trim=80 50 80 80, clip]{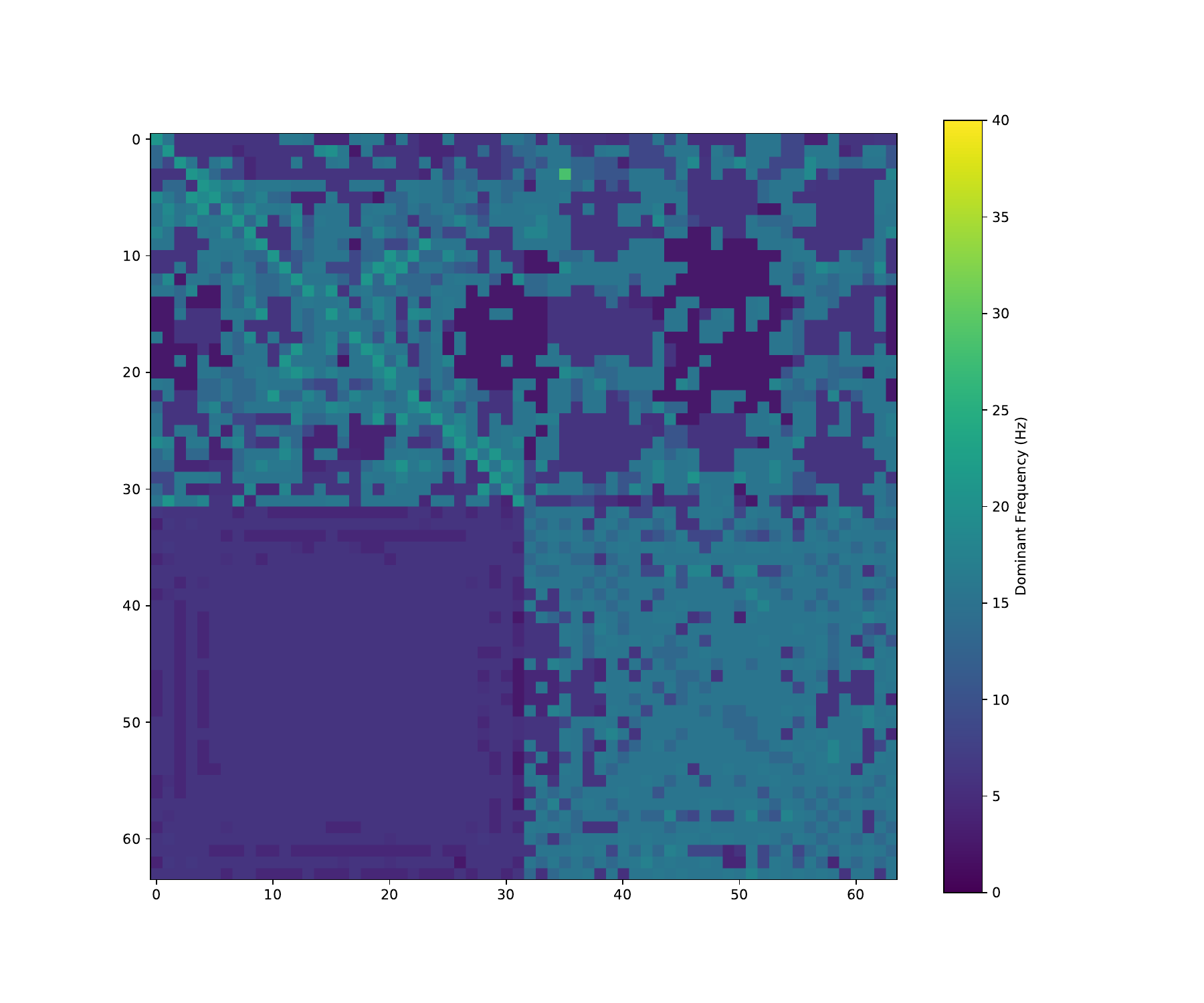}
        \subcaption{}
        \label{fig:spatial_response_e_app}
    \end{minipage}%
    \begin{minipage}{0.24\textwidth}
        \centering
        \includegraphics[width=\textwidth, trim=80 40 80 80, clip]{figures/white_noise_various_freq/frequency_spectrum_selected_points_20Hz.pdf}
        \subcaption{}
        \label{fig:spatial_response_f_app}
    \end{minipage}%
    \begin{minipage}{0.24\textwidth}
        \centering
        \includegraphics[width=\textwidth, trim=80 40 80 80, clip]{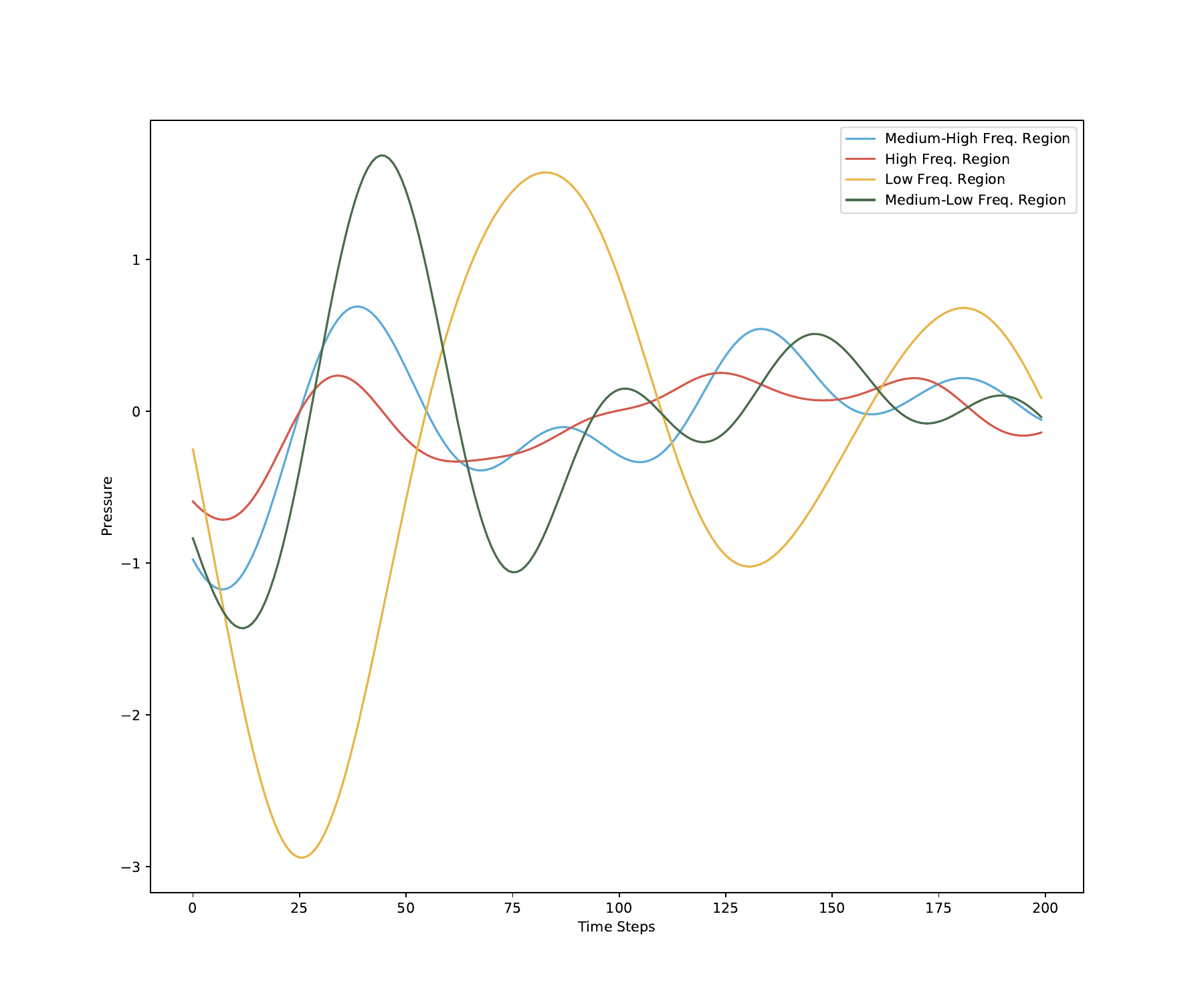}
        \subcaption{}
        \label{fig:spatial_response_g_app}
    \end{minipage}%
    \begin{minipage}{0.24\textwidth}
        \centering
        \includegraphics[width=\textwidth, trim=80 50 80 80, clip]{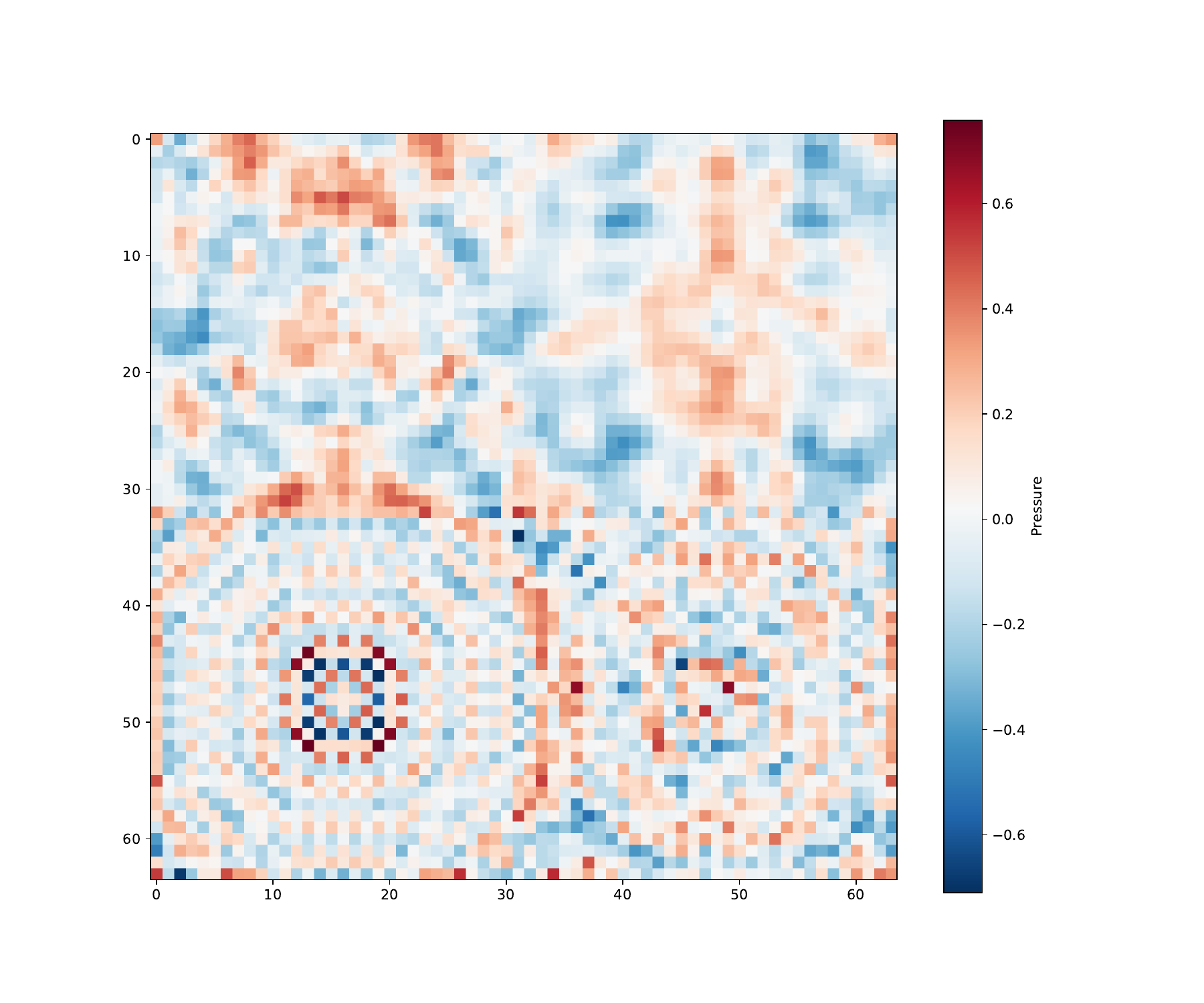}
        \subcaption{}
        \label{fig:spatial_response_h_app}
    \end{minipage}

    \vspace{0.5em} 

    \begin{minipage}{0.24\textwidth}
        \centering
        \includegraphics[width=\textwidth, trim=80 50 80 80, clip]{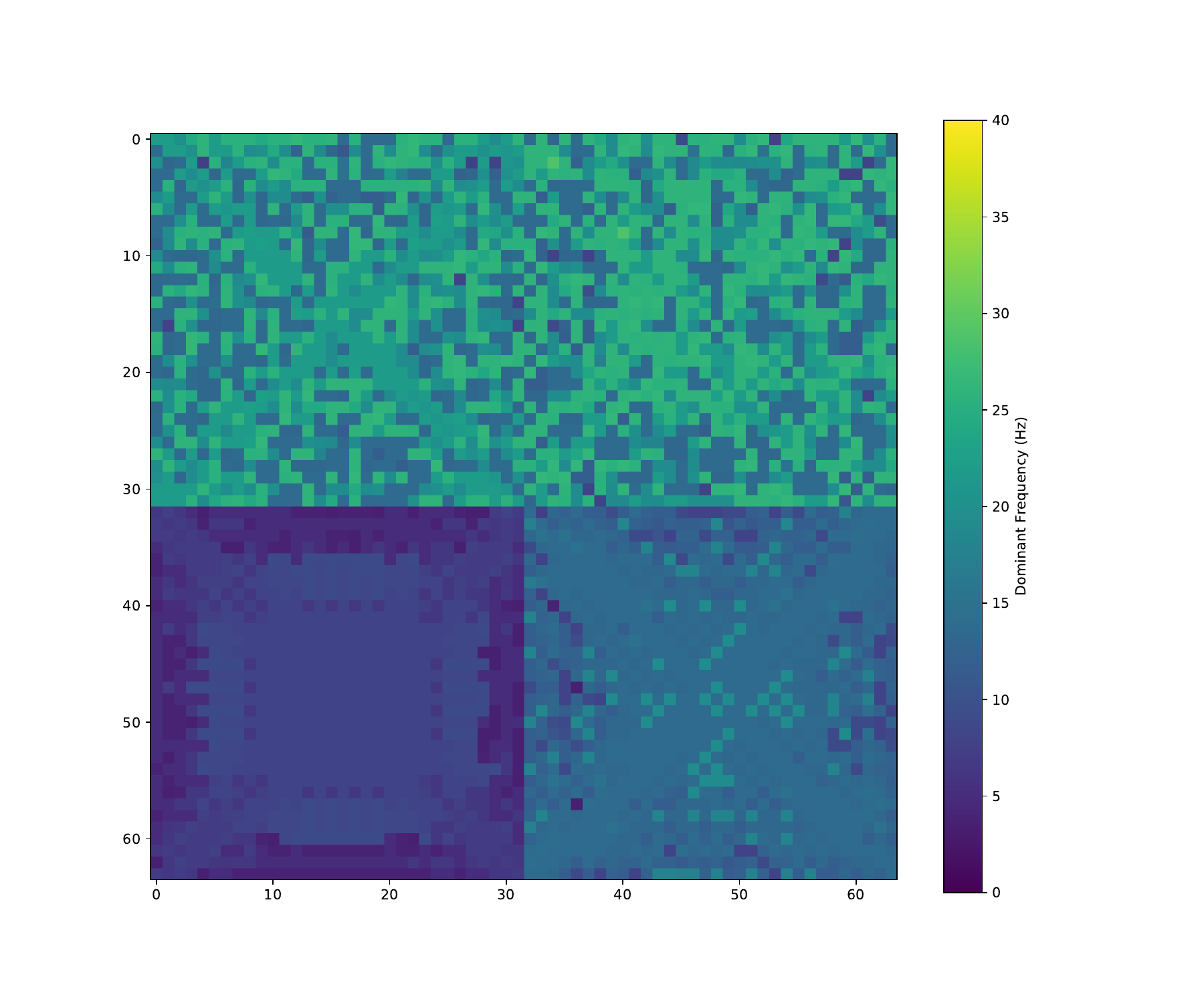}
        \subcaption{}
        \label{fig:spatial_response_i_app}
    \end{minipage}%
    \begin{minipage}{0.24\textwidth}
        \centering
        \includegraphics[width=\textwidth, trim=80 40 80 80, clip]{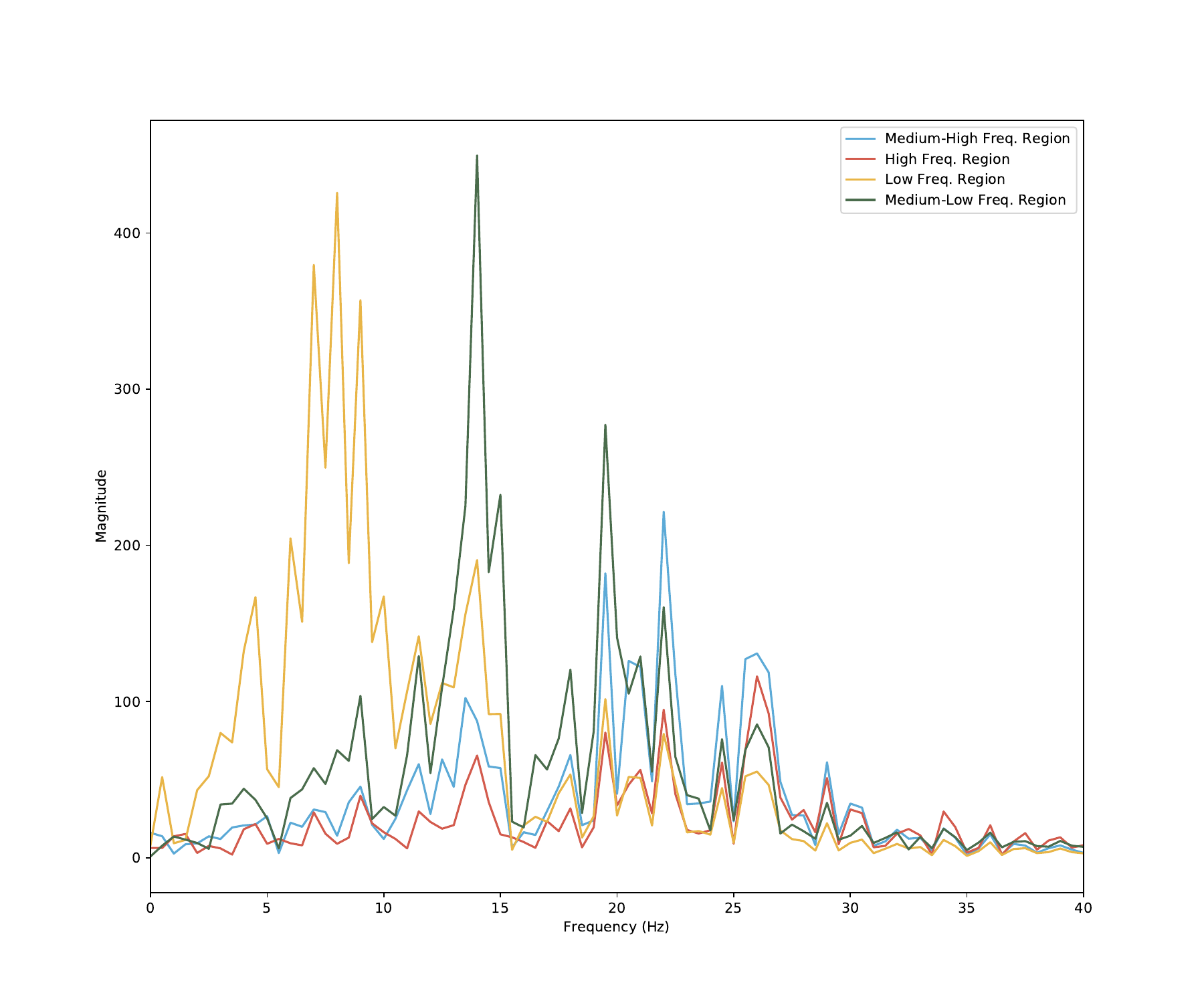}
        \subcaption{}
        \label{fig:spatial_response_j_app}
    \end{minipage}%
    \begin{minipage}{0.24\textwidth}
        \centering
        \includegraphics[width=\textwidth, trim=80 40 80 80, clip]{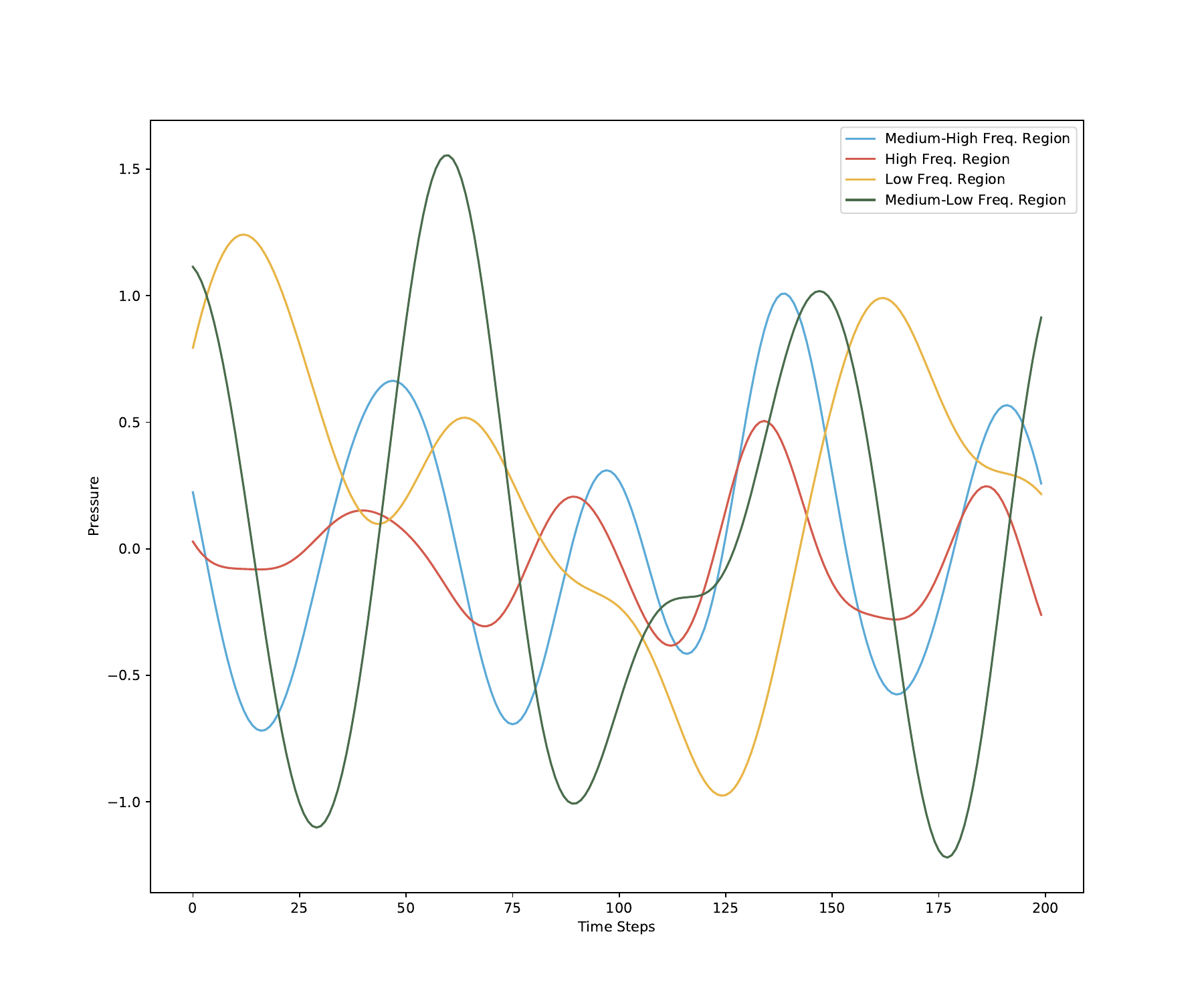}
        \subcaption{}
        \label{fig:spatial_response_k_app}
    \end{minipage}%
    \begin{minipage}{0.24\textwidth}
        \centering
        \includegraphics[width=\textwidth, trim=80 50 80 80, clip]{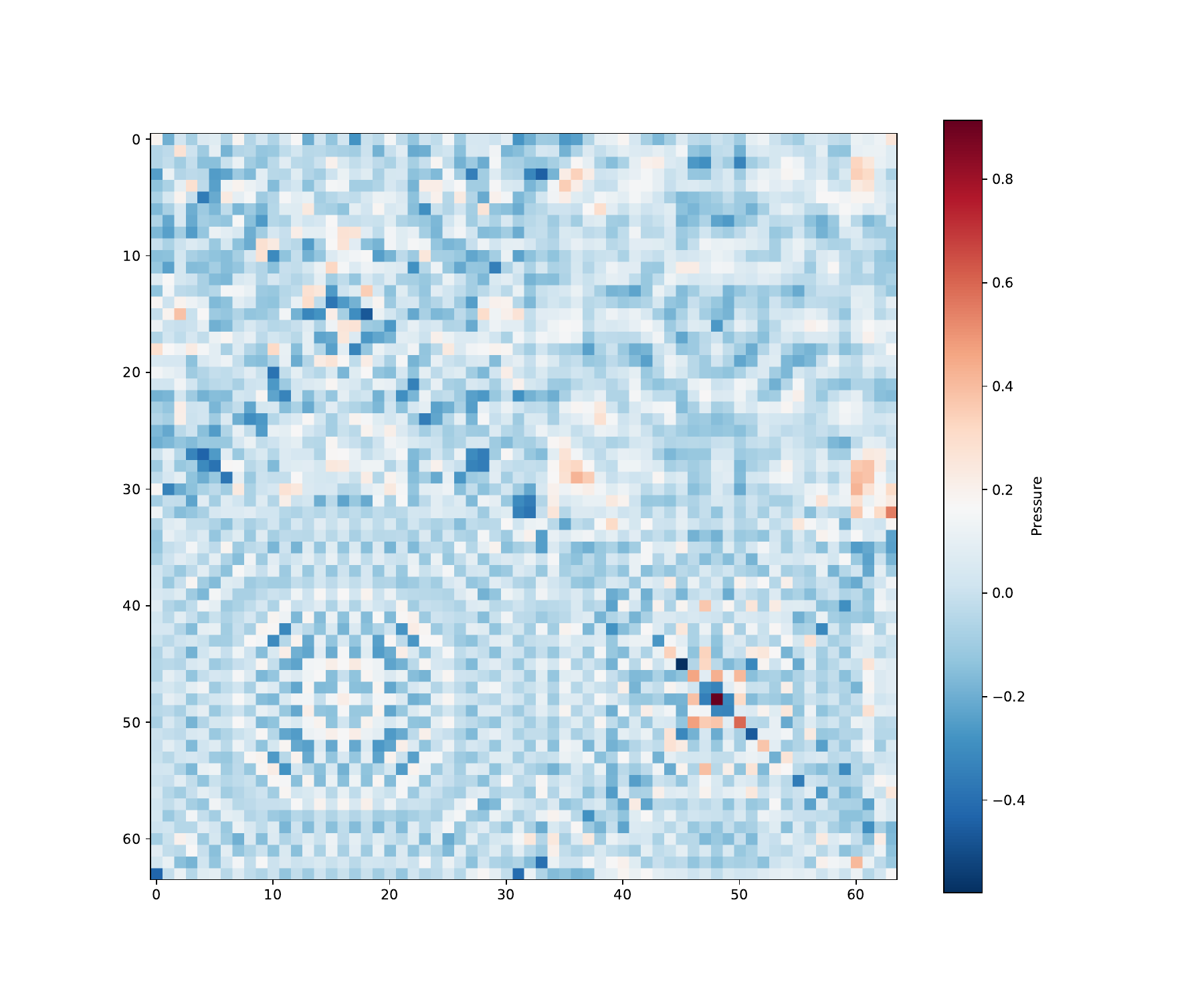}
        \subcaption{}
        \label{fig:spatial_response_l_app}
    \end{minipage}

    \caption{\textbf{Spatially partitioned frequency-selective responses of the 64 \( \times \) 64 BioOSS under varying white noise input bands.}  The three rows correspond to different input frequency bands: (a)--(d) 0–10 Hz, (e)--(h) 0–20 Hz, and (i)--(l) 0–30 Hz. Within each row: (a, e, i) depict the spatial distribution of dominant frequencies across the four quadrants of the grid, illustrating distinct regional frequency responses despite identical white noise excitation; (b, f, j) show the frequency spectra at the center of each region, highlighting region-specific resonant behavior; (c, g, k) present the time-domain responses at selected points during the final 200 time steps, reflecting the differentiated temporal dynamics induced by the frequency-selective properties of each partition; (d, h, l) display the final pressure field \(P\) after 2,000 time steps, revealing the emergent spatial patterns shaped by the respective input conditions.}
\label{fig:spatial_response_different_freq}
\end{figure}

In this experiment, a spatially partitioned 64 \(\times\) 64 BioOSS structure was stimulated with white noise inputs of varying frequency bands (0--10 Hz, 0--20 Hz, and 0--30 Hz) to investigate its frequency-selective response properties, as shown in Fig.~\ref{fig:spatial_response_main}. Each quadrant of the grid was designed to exhibit a distinct natural frequency preference. The dominant frequency response maps (Figs.~\ref{fig:spatial_response_a_app}, \ref{fig:spatial_response_e_app}, and \ref{fig:spatial_response_i_app}) reveal that each spatial region selectively responds to components of the input signal that match its predefined natural frequency. When the input white noise was limited to the 0--10 Hz band (first row), all regions exhibited oscillations concentrated around 10 Hz, as no higher frequency components were available to activate region-specific resonances. As the input bandwidth increased to 0--20 Hz (second row), the bottom quadrants began to resonate at their natural frequencies within the lower band, while the top quadrants, unable to find matching excitations at their designated higher natural frequencies, adapted by oscillating around the upper limit of the input band near 20 Hz. This behavior is clearly observable in the frequency spectra and time-domain responses at selected points (Figs.~\ref{fig:spatial_response_f_app} and \ref{fig:spatial_response_g_app}). Finally, with a 0--30Hz input (third row), the top left quadrant was also able to resonate at its natural frequency within the newly available band (Figs.~\ref{fig:spatial_response_i_app}--\ref{fig:spatial_response_k_app}), whereas the top right quadrant continued to oscillate predominantly around 30 Hz, indicating that its natural resonance lies beyond 30 Hz and thus remained driven by the highest available frequencies. The final pressure field snapshots (Figs.~\ref{fig:spatial_response_d_app}, \ref{fig:spatial_response_h_app}, and \ref{fig:spatial_response_l_app}) illustrate the emergent spatial patterns after 2,000 time steps for each input condition. These observations demonstrate the spatial-frequency selectivity of the BioOSS and its capacity to filter and respond dynamically to broadband excitations in a region-specific manner.

\section{Limitation}
\label{appendix:limit}
The model’s non-linear dynamics introduce mild training complexity and hyperparameter sensitivity. Moreover, the linear encoder limits the expression of intrinsic oscillatory behavior, making the model a partial abstraction of biological signals.

\end{document}